\documentclass[letterpaper, 10 pt, conference]{ieeeconf}  

\IEEEoverridecommandlockouts                              

\usepackage{graphics} 
\usepackage{epsfig} 
\usepackage{hyperref}
\usepackage{balance}
\usepackage{mathptmx} 
\usepackage{times} 
\usepackage{amsmath} 
\usepackage{amssymb}  
\usepackage{interval}
\usepackage{threeparttable}
\usepackage{booktabs}
\usepackage{adjustbox}
\usepackage{multicol}
\usepackage{lipsum}
\usepackage{graphicx}
\usepackage{booktabs}
\usepackage{siunitx} 
\usepackage{multirow}
\usepackage{threeparttable}
\usepackage{caption}
\usepackage{caption}
\usepackage{pgfplots}
\usepackage{pgfplotstable}
\usepackage{subcaption}
\usepackage{tikz}
\usepackage{array}
\usepackage{pgfplots}
\usepackage[dvipsnames]{xcolor}
\pgfplotsset{compat=1.18}
\usepgfplotslibrary{groupplots}
\title{\LARGE \bf
RoboLight: A Dataset with Linearly Composable Illumination \\for Robotic Manipulation}

\author{Shutong Jin$^{*}$, Jin Yang$^{*}$, Muhammad Zahid and Florian T. Pokorny
\thanks{$*$Equal contribution. The authors are with the School of Electrical Engineering and Computer Science, KTH Royal Institute of Technology
        {\tt\{shutong, jin4, mzmi, fpokorny\}@kth.se}. This work was partially supported by the Wallenberg AI, Autonomous Systems and Software Program (WASP) funded by the Knut and Alice Wallenberg Foundation. The computations were enabled by the supercomputing resource Berzelius provided by the National Supercomputer Centre at Linköping University and the Knut and Alice Wallenberg Foundation, Sweden.%
        }
}

\begin{document}

\maketitle
\thispagestyle{empty}
\pagestyle{empty}

\begin{abstract}
In this paper, we introduce \textit{RoboLight}, the first real-world robotic manipulation dataset capturing synchronized episodes under systematically varied lighting conditions.
\textit{RoboLight} consists of two components.
(a) \textit{RoboLight-Real} contains 2,800 real-world episodes collected in our custom \textit{Light Cube} setup, a calibrated system equipped with eight programmable RGB LED lights.
It includes structured illumination variation along three independently controlled dimensions: color, direction, and intensity. 
Each dimension is paired with a dedicated task featuring objects of diverse geometries and materials to induce perceptual challenges.
All image data are recorded in high-dynamic-range (HDR) format to preserve radiometric accuracy.
Leveraging the linearity of light transport, we introduce (b) \textit{RoboLight-Synthetic}, comprising 196,000 episodes synthesized through interpolation in the HDR image space of \textit{RoboLight-Real}. 
In principle, \textit{RoboLight-Synthetic} can be arbitrarily expanded by refining the interpolation granularity.
We further verify the dataset quality through qualitative analysis and real-world policy roll-outs, analyzing task difficulty, distributional diversity, and the effectiveness of synthesized data. 
We additionally demonstrate three representative use cases of the proposed dataset.
The full dataset, along with the system software and hardware design, are available at https://github.com/ShutongJIN/RoboLight.
\end{abstract}

\section{INTRODUCTION}
Data has emerged as a critical asset in robotics~\cite{amato2025data}.
Large-scale datasets~\cite{o2024open,khazatsky2024droid} support the training of foundation models~\cite{black2024pi0, kim2024openvla} and world models~\cite{agarwal2025cosmos}, by providing broad coverage of robotic interaction patterns.
At the same time, datasets function as diagnostic benchmarks~\cite{heo2023furniturebench}.
Evaluating policies across heterogeneous datasets reveals failure modes and motivates improvements in generalization~\cite{xie2024decomposing, gao2024efficient}.
These combined roles have driven significant data curation efforts in robotic manipulation, emphasizing both scale and diversity across embodiments, tasks and environments~\cite{o2024open, khazatsky2024droid}.

Despite these advances, lighting, a pervasive factor in real-world environments, remains notably underrepresented as a focus area in existing datasets.
While prior studies indicate that learned policies are highly sensitive to lighting changes~\cite{xie2024decomposing, mandi2022cacti, jin2025physically}, most of the existing datasets are collected under stable laboratory lighting with little variation.
Several factors contribute to this data gap. 
First, environmental illumination complicates data collection.
Robotic data collection often spans long time horizons, during which uncontrolled light sources, such as light shining through windows or display monitors, can introduce unintended variation.
This interference can obscure the designed lighting conditions and reduce dataset consistency and reliability.
Second, achieving dense coverage of lighting variations is costly and often impractical. 
Even with a single light source, multiple interacting factors, such as color, intensity, and direction, must be considered.
When multiple light sources are present, these factors interact combinatorially, making large-scale curation with sufficient distributional coverage difficult.
Moreover, limited reproducibility of lighting conditions constrains practical dataset reuse.
Unlike manipulation object sets~\cite{calli2015ycb} that can be reproduced through purchasing or 3D printing, lighting re-creation depends on precise spatial configuration.
As a result, even when lighting datasets are available, reproducing identical lighting conditions for evaluation is difficult, which further limits cross-group usability.

\begin{figure}[t]
    \centering
    \includegraphics[width=\linewidth]{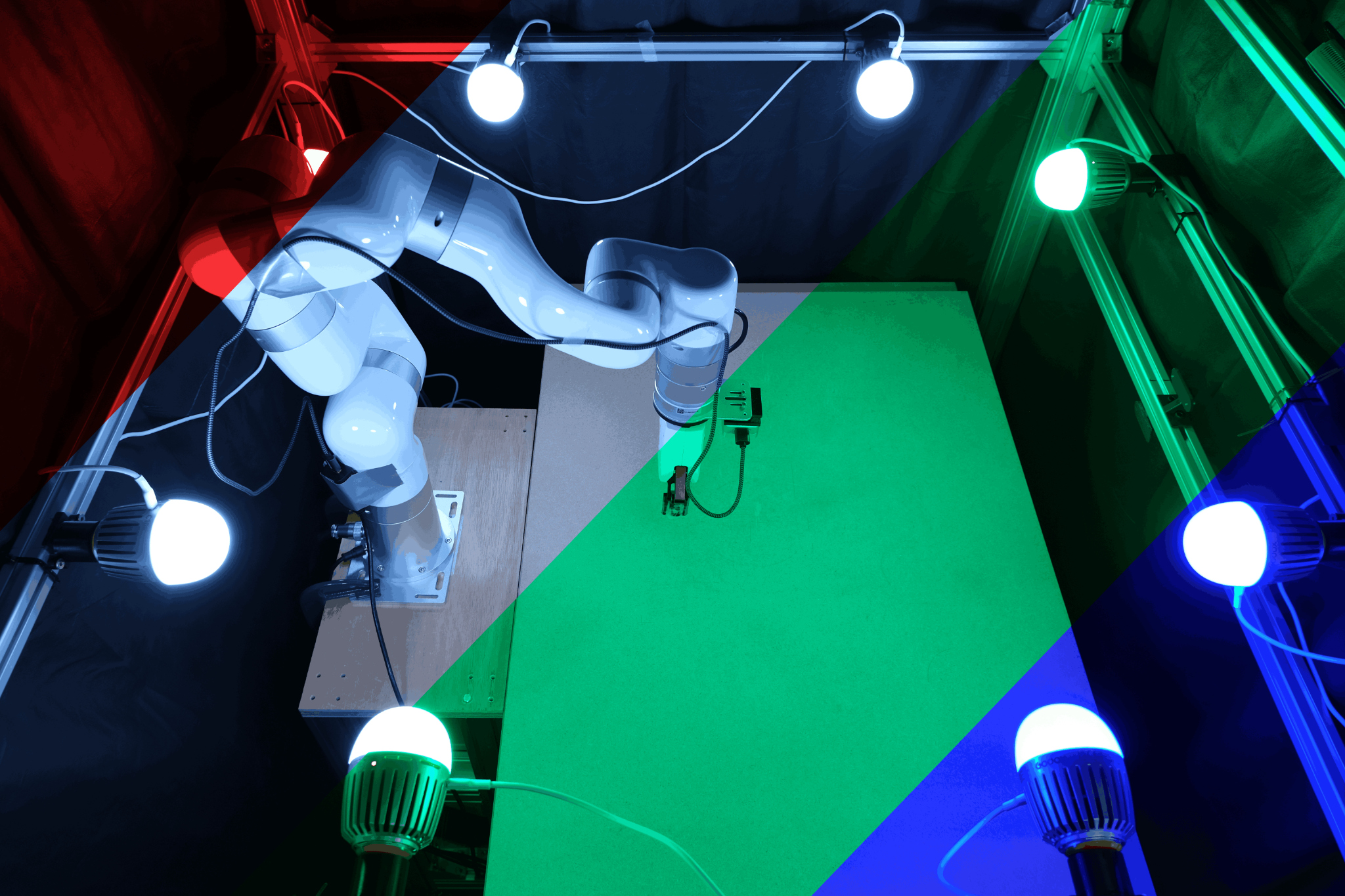}
    \caption{The \textit{Light Cube} system developed for controlled and repeatable robotic lighting data curation. We use this system to collect \textit{RoboLight}, the first real-world robotic manipulation dataset capturing synchronized episodes under systematically varied lighting conditions. The figure shows a composite overlay of four programmatically configured lighting conditions in the \textit{Light Cube}. }
    \label{fig:light_cube}
    \vspace{-1em}
\end{figure}

However, illumination remains a primary source of variability in real-world environments, changing across spatial locations and over time.
There is an urgent need for a dataset that captures such variations and enables systematic study of their impact on robotic policies.
Motivated by this, we propose \textit{RoboLight}, the first real-world robotic manipulation dataset with systematic lighting variation.
To eliminate uncontrolled external illumination and provide precisely controlled internal lighting, we construct a fully enclosed environment (\textit{Fig.}~\ref{fig:light_cube}, the \textit{Light Cube}).
By systematically controlling eight programmable RGB LED lights in the \textit{Light Cube}, we collect \textit{RoboLight-Real}, comprising 2,800 episodes with structured lighting variations.
The variations are organized along three illumination dimensions: color (red, green, blue), direction (front, rear, left, right), and intensity (140, 700, and 1400 lux). 
Additional subsets are collected under white lighting for common robotic settings and under composed lighting for data quality verification.
Each illumination dimension is paired with a task designed to expose perceptual challenges, involving 11 objects with material properties ranging from matte to highly specular.
Illumination is the only varying factor, with robot trajectories synchronized across conditions.
All images are recorded in high-dynamic-range (HDR) format to preserve radiometric accuracy.
Leveraging this linear radiometric representation and trajectory synchronization, we further synthesize \textit{RoboLight-Synthetic} via weighted interpolation in HDR space, generating 196,000 episodes. 
In principle, \textit{RoboLight-Synthetic} can be arbitrarily expanded by refining the interpolation granularity.
To conclude, our contributions are fourfold:
\begin{itemize}
\item We propose \textit{RoboLight}, the first real-world robotic manipulation dataset capturing synchronized episodes under systematically varied lighting conditions.
\item We introduce an interpolation-based data scaling strategy that synthesizes new lighting conditions via linear interpolation in HDR space, leveraging the linearity of light transport.
\item We verify the dataset quality through qualitative analysis and real-world policy roll-outs, assessing task difficulty, distributional diversity, and the effectiveness of \textit{RoboLight-Synthetic}. 
\item We demonstrate three representative use cases of the proposed dataset, ranging from lighting robustness benchmarking to HDR-enabled visual condition scaling.
\end{itemize}

\section{RELATED WORK}
\subsection{Large-Scale Datasets for Robotic Manipulation}
The robotics community has been actively accumulating large-scale datasets to support both robust single-task learning and generalist training. 
Single-task datasets typically focus on repeated task execution under object and environment variation.
A prominent example is the MIT Push dataset~\cite{yu2016more}, where object geometry and material properties are systematically varied, resulting in over one million planar pushing interactions.
Such datasets cover a broad range of manipulation skills, including pushing~\cite{yu2016more, bauza2019omnipush,jin2024physics}, grasping~\cite{yang2019replab, song2020grasping,jin2025can}, assembly~\cite{heo2023furniturebench, sliwowski2025reassemble}, pouring~\cite{zhou2023train} and etc.
Recent work has increasingly focused on datasets for generalist training, characterized by multiple embodiments, multiple tasks, and multiple environments~\cite{o2024open, khazatsky2024droid}.
One representative example is the Open X-Embodiment dataset~\cite{o2024open}, which covers 22 embodiments and 527 tasks.
Across existing real-world datasets, lighting is typically limited to stable indoor illumination, with fluctuations treated as real-world noise rather than an explicit variable.
Among the few datasets that explicitly consider lighting variation~\cite{yang2019replab, song2020grasping}, illumination is grouped into broad ``in-the-wild" categories such as indoor and window lighting. 
These variations can verify policy robustness; however, they are difficult to reproduce, and uncontrolled natural light undermines dataset consistency, limiting their suitability for systematic analysis.
Lighting has also been studied in simulation~\cite{xie2024decomposing,tobin2017domain}, yet achieving high-fidelity rendering remains challenging, as accurate light transport requires precise geometry and calibrated material properties that are rarely available in robotic simulation assets~\cite{zhou2025olatverse}.
In this paper, we present \textit{RoboLight}, the first real-world robotic manipulation dataset that treats illumination as a primary variation axis, with systematic and repeatable variation in color, direction, and intensity.

\subsection{Illumination Studies in Computer Vision and Graphics}
Two types of lighting have been widely studied in computer vision and graphics: low-light conditions and directional lighting.
Low-light conditions are commonly incorporated into datasets to assess model robustness, given their challenges for vision-centric tasks.
Representative low-light datasets cover applications such as object detection~\cite{loh2019getting}, segmentation~\cite{sun2019see}, autonomous driving~\cite{lin2024study}, face recognition~\cite{beveridge2010quantifying}, and hand gesture recognition~\cite{haroon2022hand}.
Another major line of work focuses on directional lighting~\cite{zhou2025olatverse, liu2023openillumination}. 
In this setting, images are typically captured in dark environments while a single point light source illuminates the subject from different directions. 
The Yale Face Database~\cite{lee2005acquiring} and the CMU Pose, Illumination, and Expression dataset~\cite{sim2002cmu} are two prominent examples, where human faces are photographed under individually controlled lighting directions in an enclosed dark environment.
This type of design has prompted investigations into lighting composition, exploiting the linearity of light transport to model human faces under a wide range of illumination conditions~\cite{belhumeur1998set, ramamoorthi2001efficient}.
A notable contribution is the Light Stage system~\cite{debevec2000acquiring}, which employs thousands of directional light sources to enable arbitrary lighting synthesis and has been widely adopted in the film industry~\cite{debevec2012light}.
Motivated by these approaches, we construct the \textit{Light Cube} with eight directional lights to investigate illumination effects for robotic manipulation.

\section{Preliminaries}
\subsection{HDR and LDR Images}
An image results from the interaction between illumination and scene components, as recorded by a camera. 
High-dynamic-range (HDR) and low-dynamic-range (LDR) images differ in how they represent the recorded illumination.
HDR images store pixel intensities in a linear, radiometrically calibrated space that is proportional to scene radiance, typically with 16-bit precision or higher. 
Common HDR formats include OpenEXR (.exr) and Radiance (.hdr).
LDR images encode pixel intensities in a gamma-corrected, compressed representation, usually with 8-bit precision per channel. 
Common LDR formats include JPEG (.jpg) and PNG (.png).
Given their differences, HDR images are typically used in scenarios where accurate illumination modeling is required, such as scene relighting and photometric reconstruction, whereas LDR images are primarily suited for visualization and display due to their compressed representation.
An HDR image may be mapped to an LDR image, but the inverse conversion is not possible because radiometric information is irreversibly lost.

\subsection{Linearity of Light Transport}
Light transport describes how light travels through a scene and forms an image. 
Let $L$ denote the illumination, a HDR image $I$ can be expressed as:
\begin{equation}
    I = \mathcal{T}(L).
    \label{eq:hdri}
\end{equation}
where $\mathcal{T}$ is the light transport operator determined by the scene geometry and material properties, and can be instantiated using different reflectance models~\cite{kajiya1986rendering}.

The linearity of light transport indicates that the resulting image intensity scales linearly with emitted or incident radiance~\cite{busbridge1960mathematics}.
As a result, illumination from multiple light sources combines linearly, and the corresponding HDR images can be obtained by superposing their individual contributions.
Formally, let $L_1$ and $L_2$ denote the illumination from two distinct light sources, according to the linearity of light transport, the following satisfies:
\begin{equation}
    \mathcal{T}(L_1 + L_2)= \mathcal{T}(L_1)+\mathcal{T}(L_2),
\end{equation}
Equivalently, in HDR image form,
\begin{equation}
    I_{1+2} = I_1 + I_2,
    \quad \text{where } I_i = \mathcal{T}(L_i).
    \label{eq:linearity}
\end{equation}
This behavior has been experimentally validated in real-world measurements~\cite{debevec2000acquiring, haeberli1992synthetic}.

\section{The \textit{Light Cube} System}

\subsection{Hardware Construction}
\label{sec:hardware_construction}
\begin{figure*}[t]
    \centering
    \begin{minipage}[c]{0.24\linewidth}
        \centering
        \includegraphics[width=\linewidth]{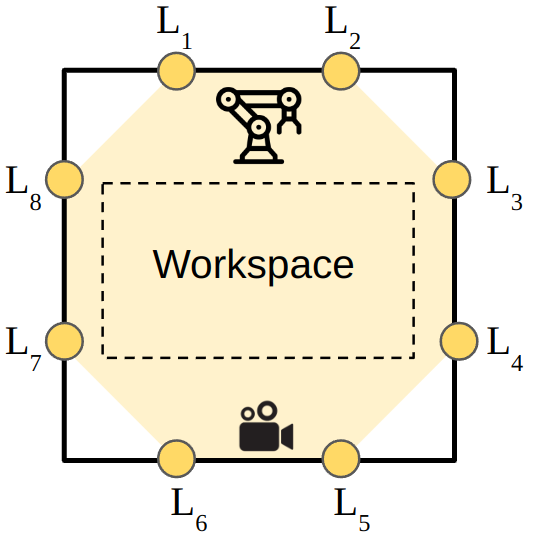}
        \vspace{-2.4em}
        \subcaption{}
        \label{fig:linear_light_real_BEV}
    \end{minipage}
    \begin{minipage}[c]{0.68\linewidth}
        \centering
        \begin{subfigure}{\linewidth}
            \centering
            \includegraphics[width=0.633\linewidth]{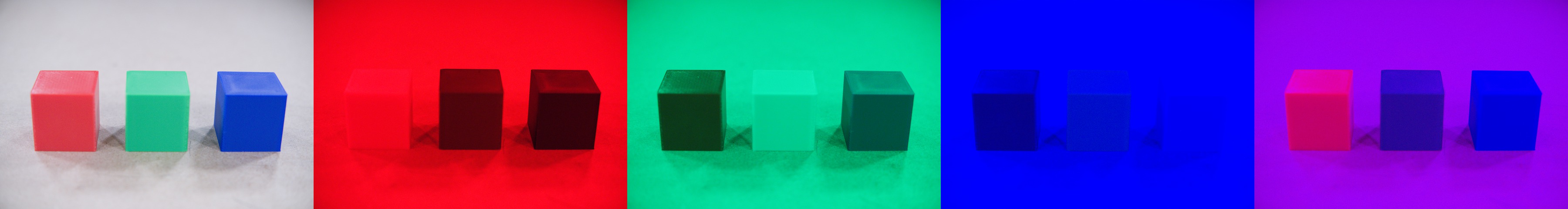}
            \vspace{-0.4em}
            \subcaption{}
            \label{fig:linear_light_real_color}
        \end{subfigure}\\[0em]
        \begin{subfigure}{\linewidth}
            \centering
            \includegraphics[width=0.8\linewidth]{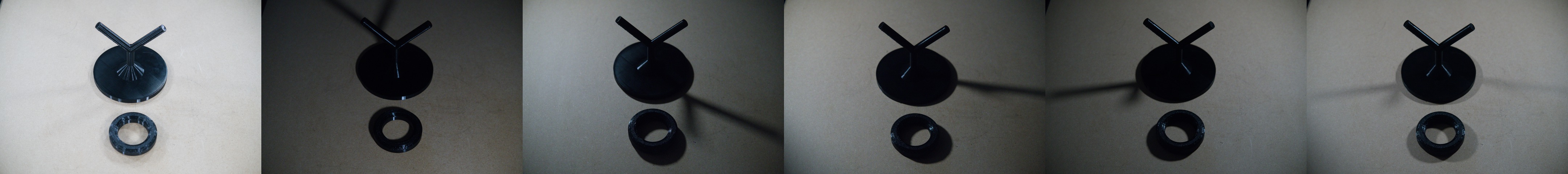}
            \vspace{-0.4em}
            \subcaption{}
            \label{fig:linear_light_real_direction}
        \end{subfigure}\\[0em]
        \begin{subfigure}{\linewidth}
            \centering
            \includegraphics[width=0.4\linewidth]{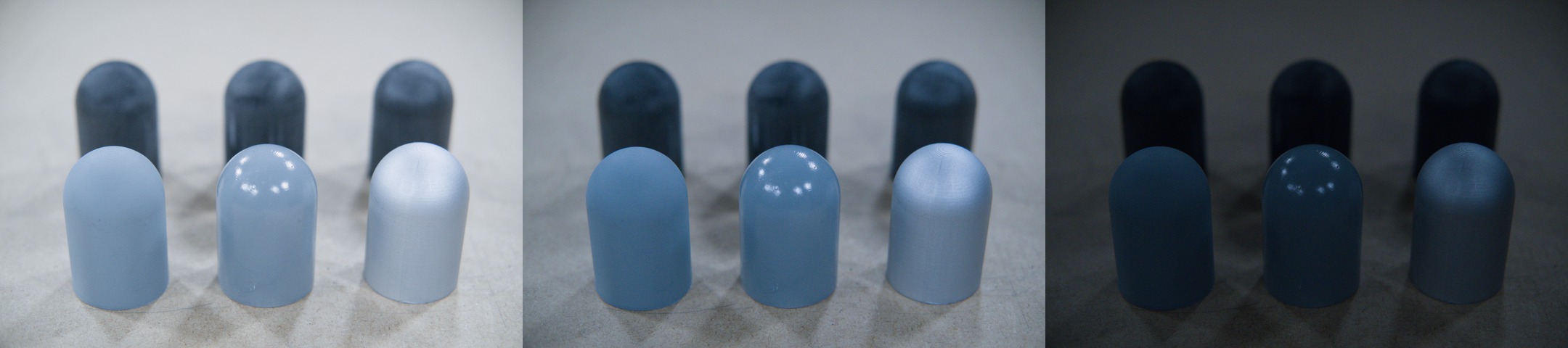}
            \vspace{-0.4em}
            \subcaption{}
            \label{fig:linear_light_real_intensity}
        \end{subfigure}
    \end{minipage}
    \vspace{-0.5em}
    \caption{(a) Top-down schematic of the \textit{Light Cube}. Overview of the \textit{RoboLight-Real} dataset: (b) the color subset, with lighting conditions from left to right: white, red, green, blue, and purple; (c) the direction subset, including all, front, rear, left, right, and left+right directional illumination; and (d) the intensity subset, featuring lighting levels of 1400 lux, 700 lux, and 140 lux. All images are cropped and zoomed in on the object sets for clearer visualization.}
    \label{fig:linear_light_real}
    \vspace{-2em}
\end{figure*}

The real-world view and top-down schematic of the \textit{Light Cube} are shown in \textit{Fig.}~\ref{fig:light_cube} and \textit{Fig.}~\ref{fig:linear_light_real_BEV}, respectively.
The system consists of five components: an enclosure, a programmable RGB LED lighting system, a workspace table, a camera system, and a robot platform.
The enclosure covers the surroundings and ceiling with light-blocking curtains to eliminate external illumination and maintain dataset consistency and experimental reproducibility.
The main frame is built from aluminum extrusion profiles, with overall dimensions of 1980 mm $\times$ 1120 mm $\times$ 1120 mm. 
These dimensions are selected based on lighting intensity simulations conducted in Blender~\cite{blender} to guarantee adequate workspace coverage.
The lighting system consists of eight Godox C10R RGB LED bulbs, connected via Bluetooth and individually controllable in both color and intensity. 
Each bulb supports a $255^3$ RGB color space, with intensity ranging from 0 lux to 320 lux as measured at the geometric center of the cube.
The workspace consists of a fixed 1000 mm × 720 mm table with height 850 mm, centered within the \textit{Light Cube}.
The camera system includes two viewpoints: a side-mounted Intel RealSense D435i ($\text{Cam}_\text{Top}$) that captures a static external view for HDR image acquisition, and a wrist-mounted Intel RealSense D405 ($\text{Cam}_\text{Wrist}$) that provides accurate close-range depth (70-500 mm).
The robotic platform consists of a 7-DoF UFactory xArm7 equipped with a force-torque sensor, mounted on a rigid base that is aligned with the workspace.

\subsection{HDR Image Processing Pipeline}
\label{sec:hdri_pipeline}
\begin{figure}[h]
    \vspace{-0.5em}
    \centering
    \begin{subfigure}{0.32\linewidth}
        \centering
        \includegraphics[width=\linewidth]{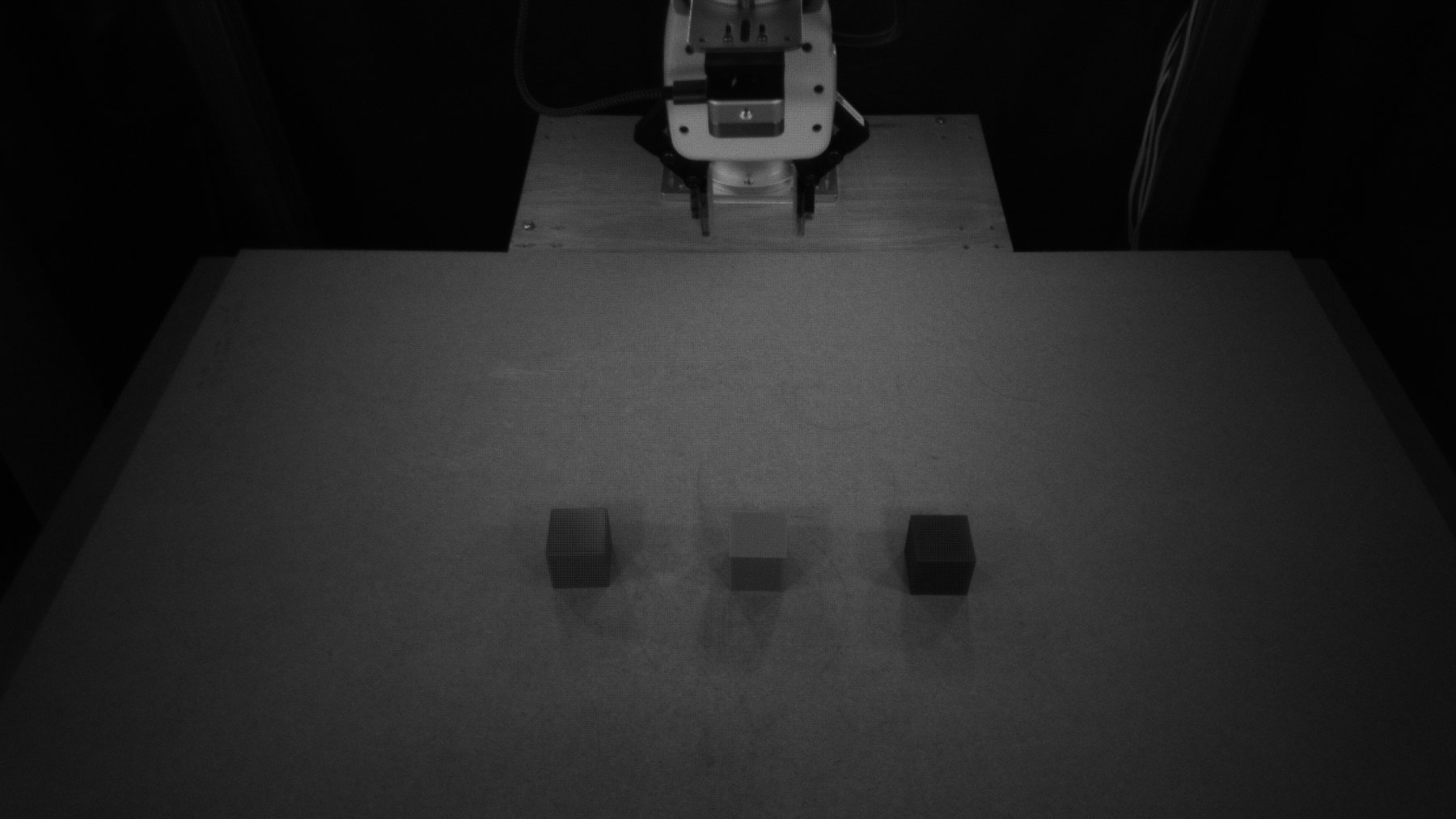}
        \vspace{-1.5em}
        \caption{}
    \end{subfigure}
    \begin{subfigure}{0.32\linewidth}
        \centering
        \includegraphics[width=\linewidth]{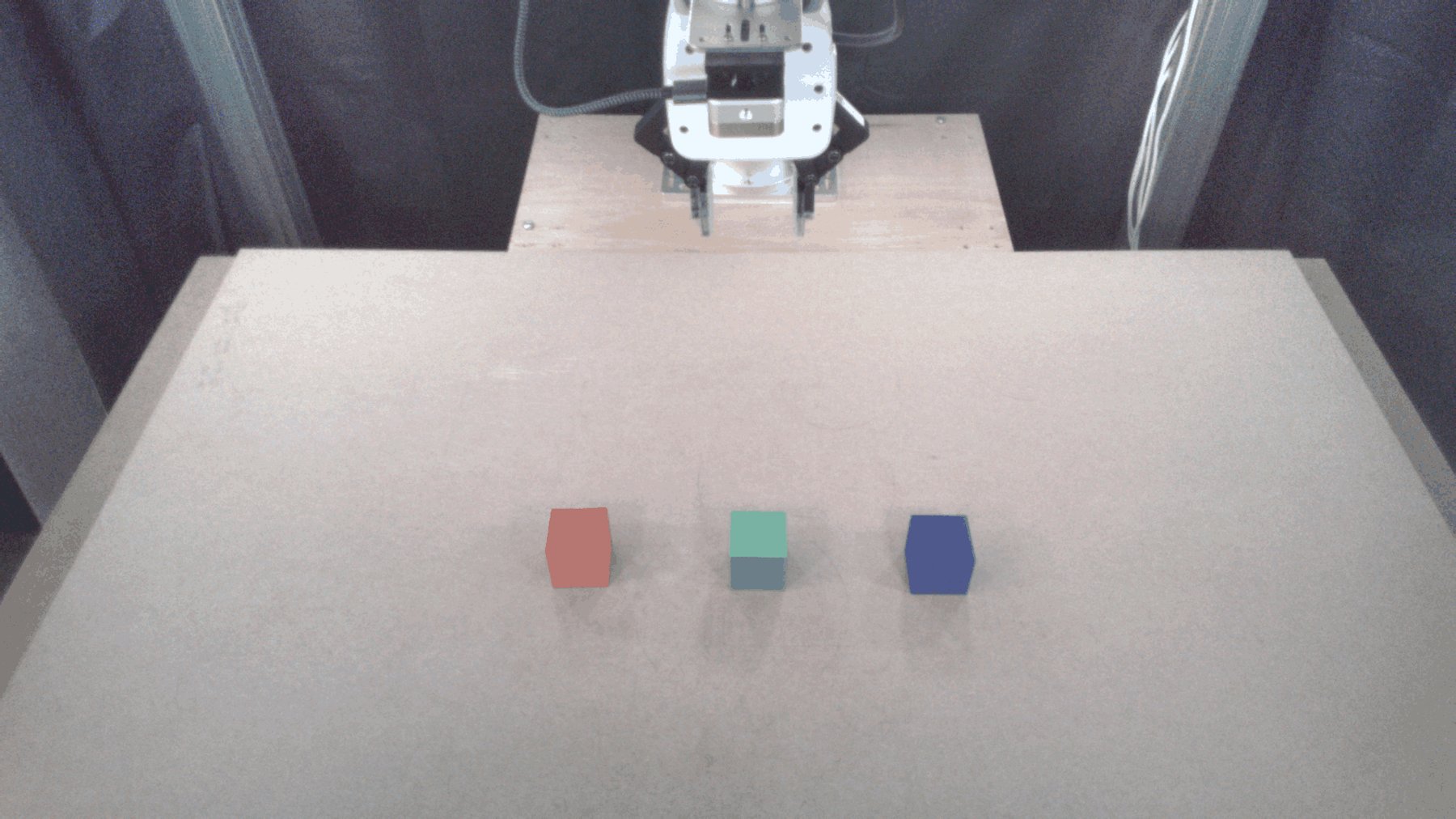}
        \vspace{-1.5em}
        \caption{}
    \end{subfigure}
    \begin{subfigure}{0.32\linewidth}
        \centering
        \includegraphics[width=\linewidth]{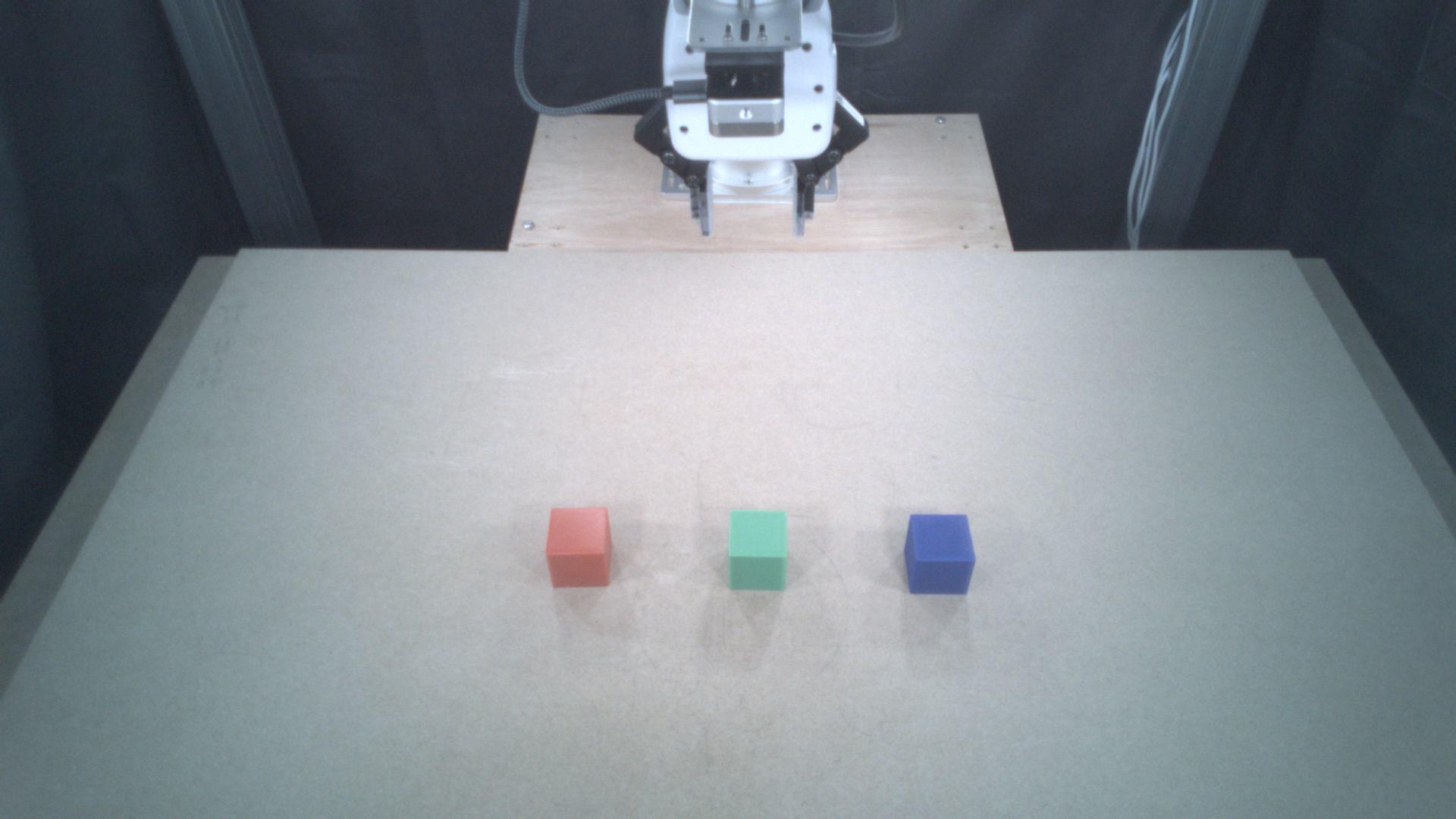}
        \vspace{-1.5em}
        \caption{}
    \end{subfigure}
    \vspace{0.5em}
    \begin{subfigure}{0.32\linewidth}
        \centering
        \includegraphics[width=\linewidth]{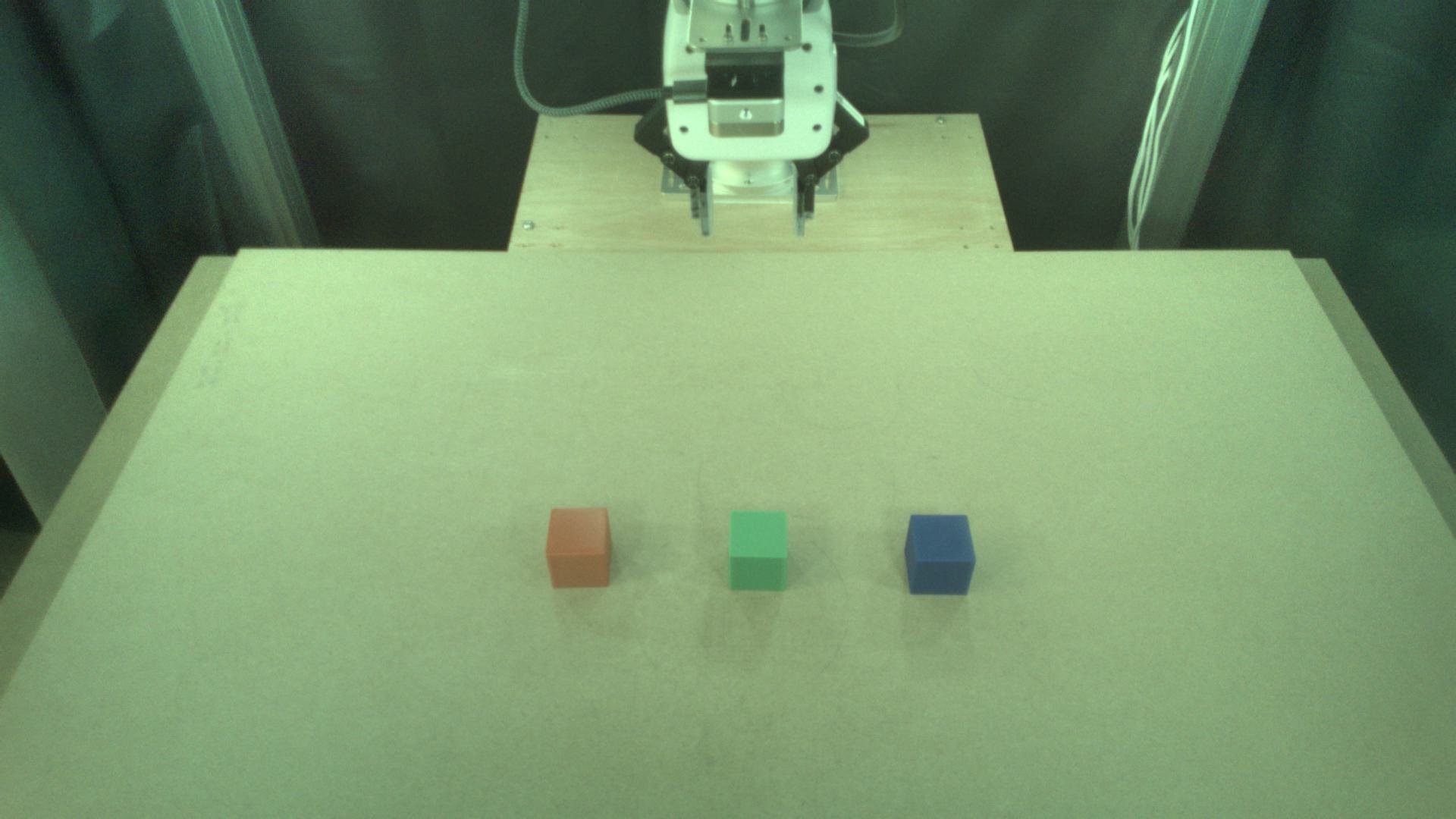}
        \vspace{-1.5em}
        \caption{}
    \end{subfigure}
    \begin{subfigure}{0.32\linewidth}
        \centering
        \includegraphics[width=\linewidth]{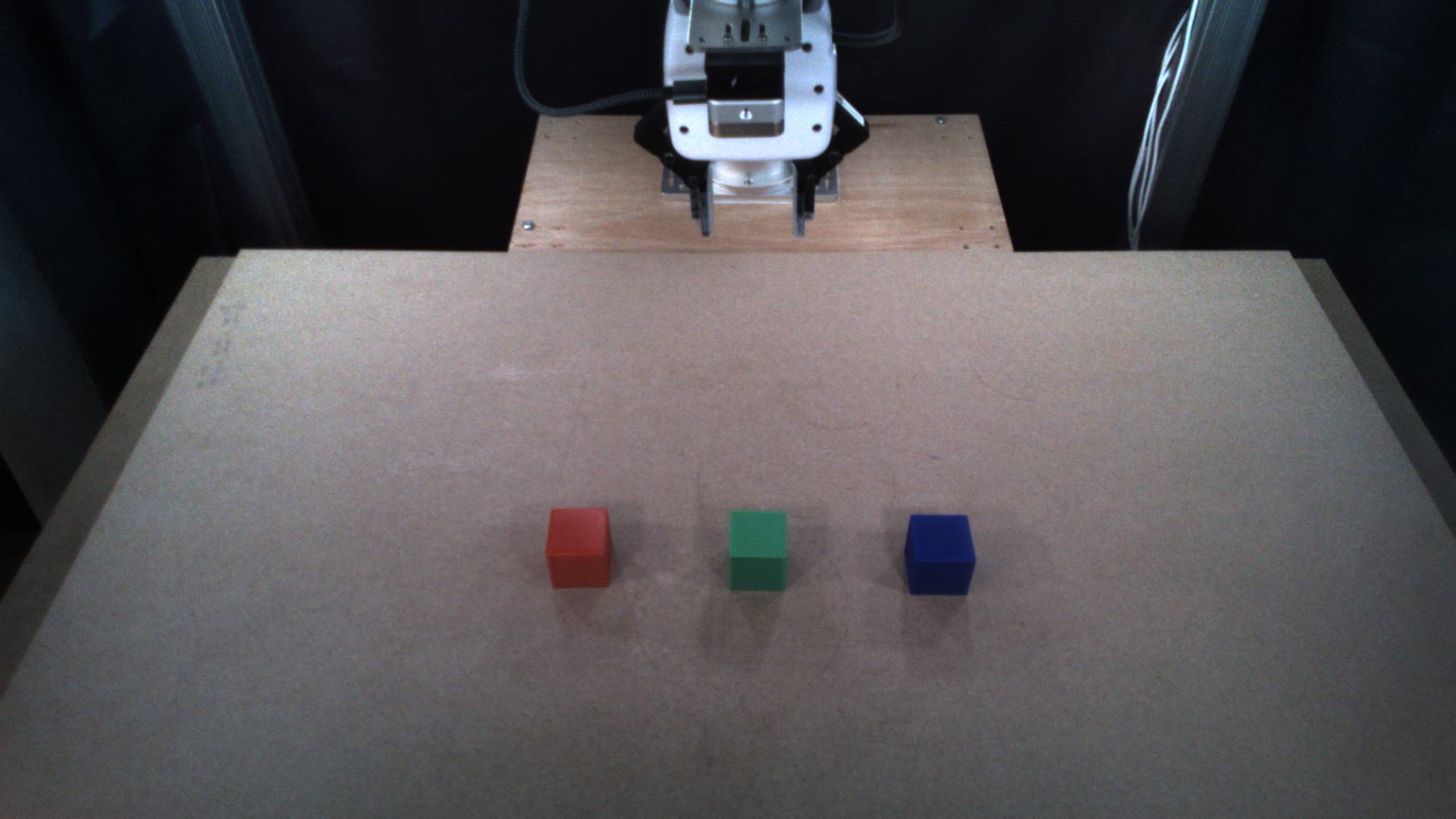}
        \vspace{-1.5em}
        \caption{}
    \end{subfigure}
    \begin{subfigure}{0.32\linewidth}
        \centering
        \includegraphics[width=\linewidth]{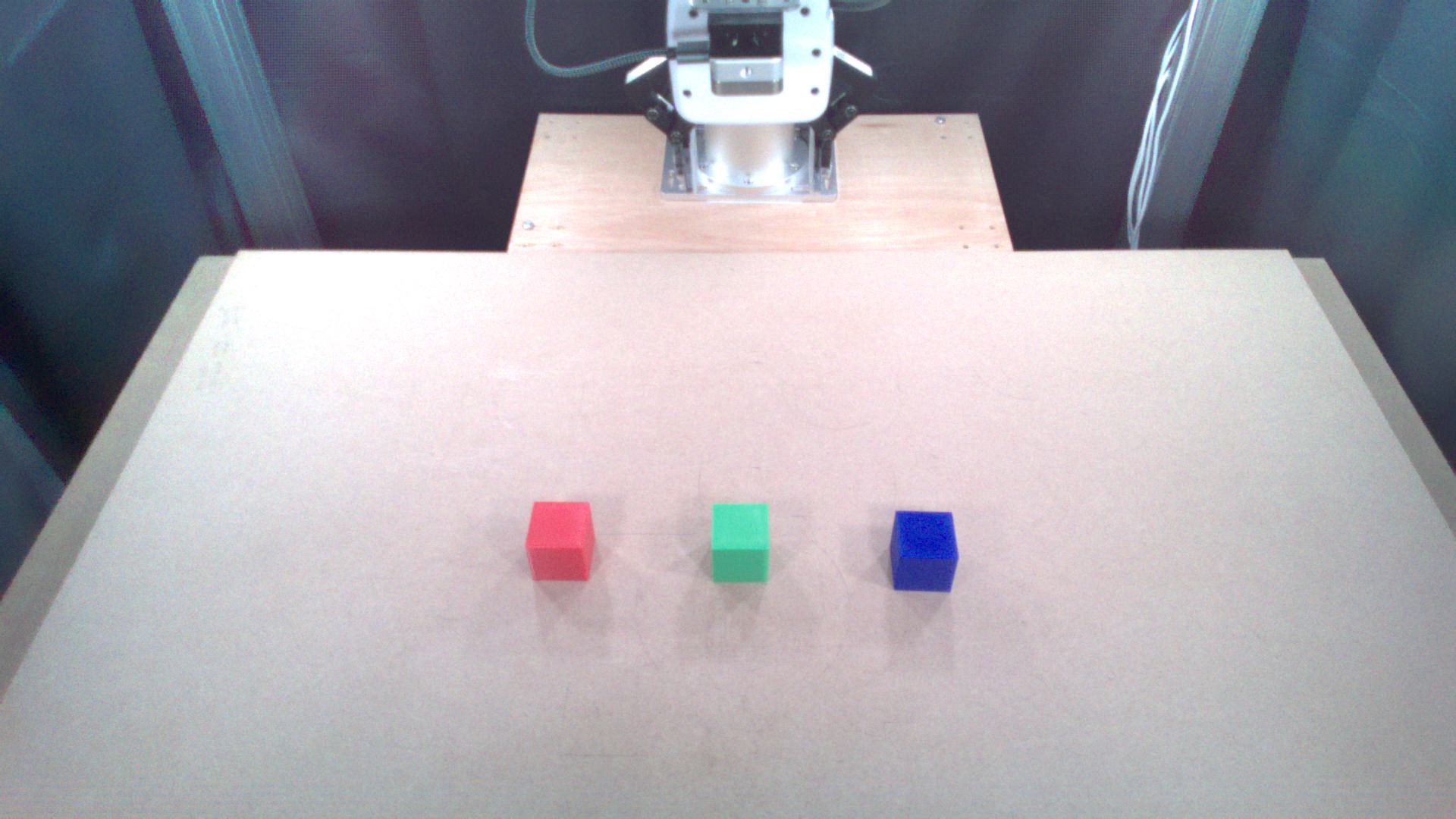}
        \vspace{-1.5em}
        \caption{}
    \end{subfigure}
    \vspace{-1em}
    \caption{Overview of the HDR image processing pipeline. (a) Unprocessed RAW16 HDR image. (b-e) Effects of removing individual processing steps: bilateral denoising, lens shading correction, white balance correction, and color and gamma correction. (f) Final PNG output after full HDR processing, used for policy training.}
    \label{fig:hdri_pipeline}
    \vspace{-1em}
\end{figure}

HDR formats are commonly supported by professional photography cameras. 
However, such cameras are often incompatible with typical robotic RGB-D setups and are costly to deploy at scale. 
In this section, we describe a custom pipeline for deriving HDR images using the widely accessible Intel RealSense D435i:
\begin{enumerate}
    \item \textit{RAW16 acquisition}: Enable uncompressed 16-bit RAW output on the RealSense camera using compatible firmware to preserve linear radiometric measurements. RAW16 serves as the HDR image format for RealSense cameras, and HDR interpolation is performed in this space.
    \item \textit{Bilateral denoising}: Apply edge-preserving bilateral filtering to suppress sensor noise while maintaining structural details.
    \item \textit{Lens shading correction}: Correct lens-induced spatial brightness attenuation.
    \item \textit{White balance correction}: Normalize color responses under varying lighting conditions.
    \item \textit{Color and gamma correction}: Map sensor responses to a calibrated color space.
    \item \textit{PNG compression}: Encode processed images for visualization and policy training.
\end{enumerate}
A visualization of the processing pipeline is provided in \textit{Fig.}~\ref{fig:hdri_pipeline}. 
This pipeline is custom-developed based on established HDR imaging and radiometric calibration procedures~\cite{reinhard2006high} and is not included in the official RealSense software development kit (SDK). 
Parameters for steps (5) and (6) are determined through the calibration procedure described in Sec.~\ref{sec:calibration}.

\subsection{Data Collection Protocol}
\label{sec:data_collection}
To collect synchronized episodes under varying lighting conditions, we employ a record-replay-reset protocol. 
A human demonstrator first provides a kinesthetic demonstration, where the trajectory is recorded. 
The robot then replays the same trajectory multiple times under different lighting conditions until all target configurations are covered.
At the end of each replay session, the robot resets the scene by executing the recorded trajectory in reverse.
We perform dataset synchronization checks after each trajectory has been replayed under all lighting conditions.

\subsection{Calibration}
\label{sec:calibration}
Calibration of the lighting system and image processing pipeline is performed before each data collection session.
As shown in \textit{Fig.}~\ref{fig:linear_light_real_BEV}, the lights are positioned in an octagonal configuration to ensure even angular coverage and to support structured lighting composition. Illumination intensity, measured in lux at the geometric center of the Light Cube, is averaged from four directions (front, rear, left, and right). 
Each measurement is repeated three times and documented to ensure consistent lighting conditions within each subset.
For camera calibration, intrinsic parameters and lens distortion coefficients are estimated using a standard checkerboard-based procedure implemented in OpenCV~\cite{zhang2002flexible}. 
Depth calibration follows the standard procedure provided in the Intel RealSense SDK.
For HDR image processing, we follow a standard color calibration pipeline using a color correction board to derive the color correction matrix (CCM), exposure, and white balance parameters required in \textit{Sec.}~\ref{sec:hdri_pipeline}.

\section{The \textit{RoboLight} Dataset}
\subsection{Tasks}
\label{sec:tasks}
\paragraph{RGB Stacking}
The robot picks up three randomly scattered cubes and stack them in a fixed order: red at the bottom, green, and blue on the top.
This task is designed to amplify the perceptual effects of lighting color variation.
As shown in \textit{Fig.}~\ref{fig:linear_light_real_color}, object appearance changes significantly under different lighting colors and may become nearly imperceptible when illumination matches object color.

\paragraph{Donut Hanging}
The robot picks up a randomly placed donut-shaped object and hangs it on a Y-shaped tree, with success defined as hooking the donut onto the nearest branch.
This task is designed to amplify the perceptual effects of variations in lighting direction.
As shown in \textit{Fig.}~\ref{fig:linear_light_real_direction}, changes in lighting direction produce pronounced shadows from both the donut and the Y-shaped structure, which can significantly affect grasping and placement. 
Improper handling of shadow-induced perception errors can lead to suboptimal grasp points and incorrect hanging trajectories.
All objects in this task are dark-colored, minimizing the influence of lighting intensity and concentrating the analysis on directional lighting variations.

\paragraph{Sparkling Sorting}
The robot picks up three silver target objects coated with distinct surface materials (matte, glossy, and metallic) into a basket in the presence of three matte black distractors.
This task is designed to amplify the perceptual effects of variations in lighting intensity.
As shown in \textit{Fig.}~\ref{fig:linear_light_real_intensity}, the target objects exhibit strong variations in reflectance due to their different surface finishes, whereas the matte distractors maintain a nearly constant appearance under varying lighting intensity.

\begin{figure*}[t]
    \centering
    \begin{subfigure}{\linewidth}
        \centering
        \includegraphics[width=\linewidth]{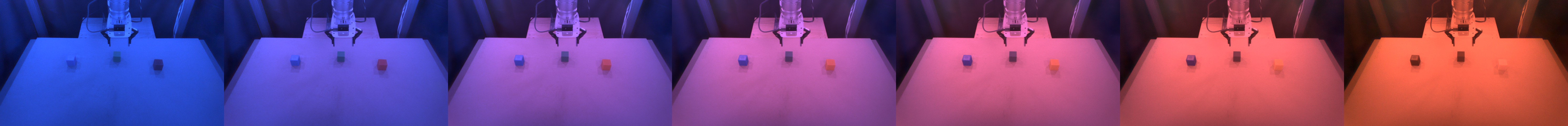}
    \end{subfigure}
    \\[0.2em]
    \begin{subfigure}{\linewidth}
        \centering
        \includegraphics[width=\linewidth]{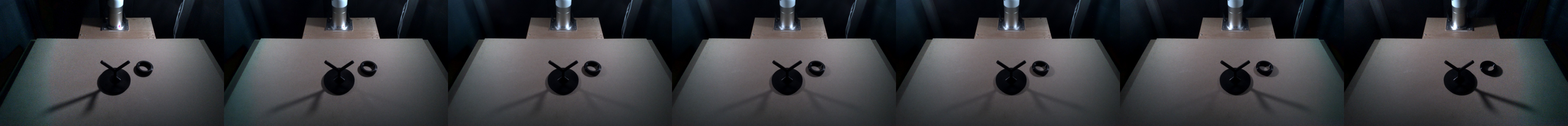}
    \end{subfigure}
    \\[0.2em]
    \begin{subfigure}{\linewidth}
        \centering
        \includegraphics[width=\linewidth]{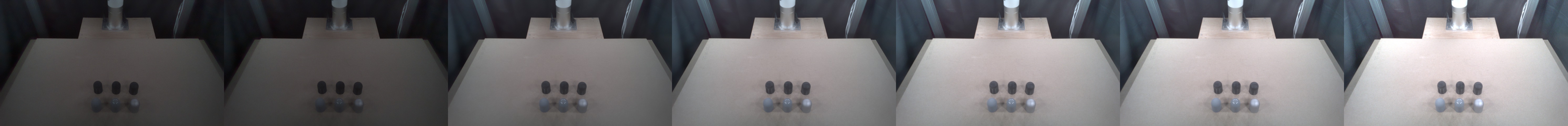}
    \end{subfigure}
\begin{tikzpicture}
    \node[left] at (2,0) {\textbf{Real}};
    \node[right] at (16,0) {\textbf{Real}};
    \draw[<->, line width=1pt] (2.7,0) -- (15.4,0)
        node[midway, above] {\textbf{Synthesized}};
    \draw[line width=1pt] (2.7,-0.25) -- (2.7,0.25);
    \draw[line width=1pt] (15.4,-0.25) -- (15.4,0.25);
\end{tikzpicture}
    \caption{Examples from \textit{RoboLight-Synthetic}. New episodes are synthesized by interpolating between HDR images from \textit{RoboLight-Real}. In each row, only the first and last images are real; all intermediate images are synthesized.}
    \label{fig:linear_light_synthesized}
    \vspace{-1em}
\end{figure*}
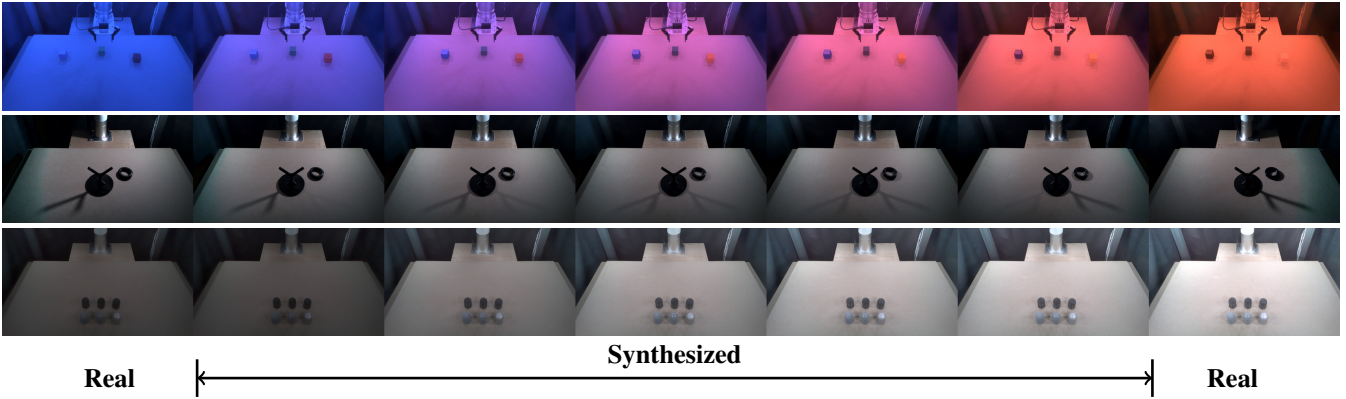

\subsection{RoboLight-Real}
As shown in \textit{Fig.}~\ref{fig:linear_light_real}, the three tasks are individually paired with systematic variation in color, direction, and intensity, yielding 14 curated subsets.
Color variation for \textit{RGB Stacking} includes the three primary components of the color space: red, green, and blue.
Direction variation for \textit{Donut Hanging} includes four principal directions: front, rear, left, and right.
Intensity variation for \textit{Sparkling Sorting} comprises two levels: 140 lux under low-light conditions and 1400 lux under common ambient lighting.
In addition, each task includes a lighting condition used as ground truth for validating \textit{RoboLight-Synthetic}, consisting of purple color, combined left and right illumination, and 700 lux intensity.
An ambient white lighting condition is also included for \textit{RGB Stacking} and \textit{Donut Hanging} to reflect common training settings.
Each subset contains 200 episodes, totaling 2,800 episodes. 
Detailed per-episode information is listed in \textit{Tab.}~\ref{tab:episode_information} and 
a summary of \textit{RoboLight-Real} is provided in \textit{Tab.}~\ref{tab:linear_light_real}

\renewcommand{\arraystretch}{1.2} 
\setlength{\tabcolsep}{8pt} 
\begin{table}[h]
\centering
\caption{Episode-level information in \textit{RoboLight-Real}.}
\label{tab:episode_information}
\begin{tabular}{llll} 
\toprule
\textbf{Modality} & \textbf{Dimension} & \textbf{Frequency} \\ 
\midrule
$\text{Cam}_\text{Top}$ RGB & $ 1920\times 1080\times3 $ & 30 Hz\\
$\text{Cam}_\text{Top}$ Depth & $ 640\times 480$ & 30 Hz\\
$\text{Cam}_\text{Wrist}$ RGB & $ 640\times480 \times3 $ & 30 Hz\\
$\text{Cam}_\text{Wrist}$ Depth& $ 640\times 480$ & 30 Hz\\
Force-torque &  $ 6$ & 10 Hz\\
Proprioception & $ 14 $ & 30 Hz\\
\bottomrule
\end{tabular}
\begin{tablenotes}
    \item Additional metadata includes object positions, lighting parameters, calibration matrices, and sensor timestamps.
\end{tablenotes}
\vspace{-1em}
\end{table}

\renewcommand{\arraystretch}{1.2} 
\setlength{\tabcolsep}{8pt} 
\begin{table}[h]
\centering
\caption{Dataset overview for \textit{RoboLight-Real}.}
\label{tab:linear_light_real}
\begin{tabular}{cccccc} 
\toprule
\textbf{Task} & \textbf{Color} &{\textbf{Direction}} & \textbf{Intensity}  \\ 
\midrule
\multirow{5}{*}{RGB Stacking}  &White  & All  & 700 lux \\
&Red & All & 172 lux\\
&Green& All & 335 lux\\
&Blue& All&72 lux\\
&Purple&All & 95 lux \\
\midrule
\multirow{6}{*}{Donut Hanging}  &White  & Four-Directional  & 330 lux \\
&White & Front & 82 lux\\
&White& Rear & 80 lux\\
&White& Left & 73 lux\\
&White& Right & 76 lux \\
&White& Left+Right & 144 lux \\
\midrule
\multirow{3}{*}{Sparkling Sorting}  &White  & All  & 140 lux \\
&White & All & 700 lux\\
&White& All & 1400 lux\\
\bottomrule
\end{tabular}
\begin{tablenotes}
\footnotesize
\item
Detailed description of color and direction settings:
``Red'' (255, 0, 0), 
``Green'' (0, 255, 0), 
``Blue'' (0, 0, 255), 
and ``Purple'' (255, 0, 255).
Using the notation in \textit{Fig.}~\ref{fig:linear_light_real_BEV},
``All'' ($\text{L}_1$, $\text{L}_2$, $\text{L}_3$, $\text{L}_4$, $\text{L}_5$, $\text{L}_6$, $\text{L}_7$, $\text{L}_8$), ``Four-Directional'' ($\text{L}_1$, $\text{L}_3$, $\text{L}_5$, $\text{L}_8$)
``Front'' ($\text{L}_5$), 
``Rear'' ($\text{L}_1$), 
``Left'' ($\text{L}_8$), 
``Right'' ($\text{L}_3$), 
and ``Left+Right'' ($\text{L}_3$, $\text{L}_8$).
\end{tablenotes}
\vspace{-1em}
\end{table}

\subsection{RoboLight-Synthetic}
\label{sec:linear_light_syn}
Based on the linearity of light transport (\textit{Eq.}~\ref{eq:linearity}), new images can be synthesized by linearly interpolating between two synchronized HDR images. 
We extend this principle to the episode level and generate new episodes from \textit{RoboLight-Real} by interpolating between synchronized episodes captured under different lighting conditions.
Let an episode be a sequence of $T$ synchronized HDR frames:
\begin{equation}
    E_1 = \{ I_1^t \}_{t=1}^{T}, 
    \quad
    E_2 = \{ I_2^t \}_{t=1}^{T}.
\end{equation}
Then a synthesized episode is defined as:
\begin{equation}
    E_{\lambda}
    =
    \lambda E_1
    +
    (1-\lambda)E_2,
    \quad \lambda \in [0,1].
\end{equation}
By discretizing $\lambda$ in steps of 0.01 between each pair of structured lighting components, we construct \textit{RoboLight-Synthetic}, comprising 196,000 episodes without additional real-world collection. 
The interpolation is performed in HDR image space using RAW16 data and is subsequently processed through the custom HDR pipeline described in Sec.~\ref{sec:hdri_pipeline} to produce PNG images for policy training.
In principle, \textit{RoboLight-Synthetic} can be arbitrarily expanded by refining the interpolation granularity.
Examples are shown in \textit{Fig.}~\ref{fig:linear_light_synthesized}.

\section{Dataset Quality Evaluation}
\subsection{Task Difficulty}
\label{sec:task_difficulty}
We assess task difficulty by training a policy from scratch and evaluating it under the same lighting conditions as in training data. 
We employ Diffusion Policy~\cite{chi2025diffusion} as a strong manipulation baseline and conduct 20 roll-outs per setting. 
Evaluation success rates are listed in \textit{Tab.}~\ref{tab:task_difficulty}.
We hypothesize that the higher difficulty of \textit{Sparkling Sorting} is due to the large number of objects and their wide spatial distribution. 
For \textit{RGB Stacking}, difficulty arises from error accumulation in the long-horizon stacking sequence, where imprecise grasps can lead to instability and failure. 
\textit{Donut Hanging} is the least difficult, likely due to its shorter execution horizon.

\renewcommand{\arraystretch}{1.2} 
\setlength{\tabcolsep}{8pt} 
\begin{table}[h]
\vspace{-1em}
\centering
\caption{Task difficulty evaluated by success rate.}
\label{tab:task_difficulty}
\begin{tabular}{lccc} 
\toprule
\textbf{RGB Stacking} & \textbf{Donut Hanging} & \textbf{Sparkling Sorting} \\ 
\midrule
0.70 & 0.95& 0.35\\
\bottomrule
\end{tabular}
\begin{tablenotes}
    \item A vanilla Diffusion Policy with a CNN backbone and default hyperparameters is used. Evaluation is performed on the \textit{RGB Stacking} (Purple), \textit{Donut Hanging} (Left+Right), and \textit{Sparkling Sorting} (700 lux).
\end{tablenotes}
\vspace{-1em}
\end{table}

\subsection{Distribution Diversity}
\label{sec:object_position_diversity}
\begin{figure}[h]
    \centering
    \includegraphics[width=0.9\linewidth]{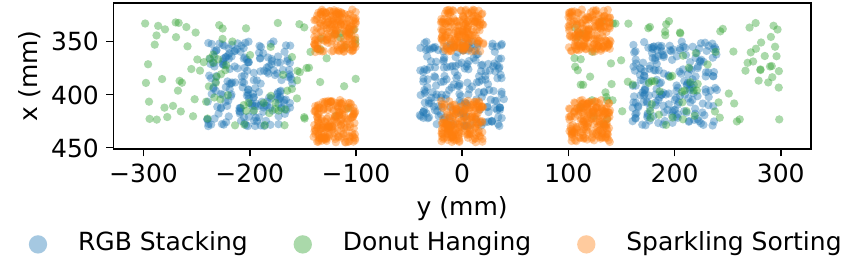}
    \vspace{-0.5em}
    \caption{Initial object position distributions for the three tasks, each computed from 200 episodes. Coordinates are in millimeters and referenced to the robot base frame. Point density varies across tasks due to differences in object count. Within each task, object positions are randomized across placement patches.}
    \label{fig:object_position_diversity}
    \vspace{-1.5em}
\end{figure}

Previous studies have shown that balanced object distributions are important for unbiased policy training~\cite{saxena2025matters}. 
To ensure this, we avoid manual resetting and instead perform robot-assisted random sampling within predefined initial placement ranges for each data collection session.
A visualization of sampled object positions across the three tasks is shown in \textit{Fig.}~\ref{fig:object_position_diversity}. The placement ranges are determined through iterative task difficulty test as described in \textit{Sec.}~\ref{sec:task_difficulty}, aiming to avoid extreme difficulty settings.

\subsection{Verification of \textit{RoboLight-Synthetic}}
\label{sec:linear_light_syn_effectiveness}
\begin{figure}[h]
\vspace{-0.5em}
\centering
  \begin{subfigure}[h]{0.9\linewidth}
    \centering
    \hspace{0.2em}\text{Ground Truth}\hspace{0.8em}\text{\textit{RoboLight-Synthetic}}\hspace{0.5em}\text{PNG-Average}\\
    \includegraphics[width=0.92\linewidth]{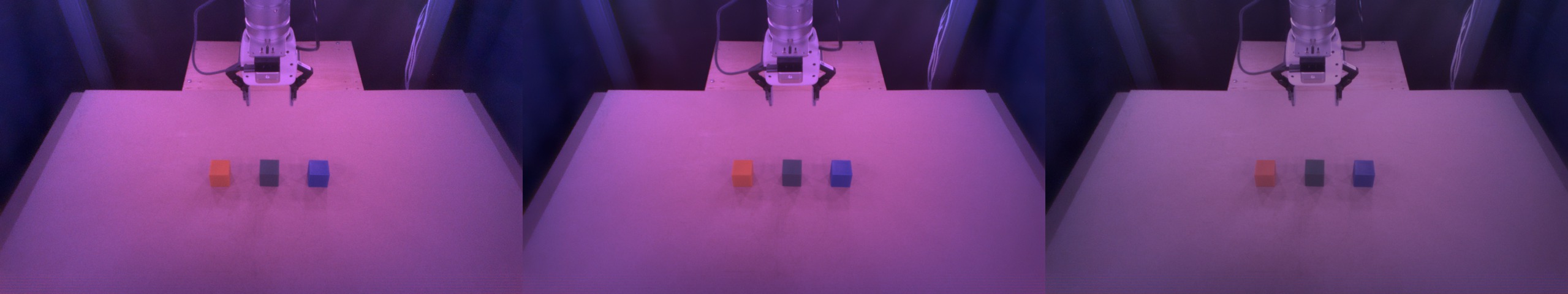}
    \includegraphics[width=\linewidth]{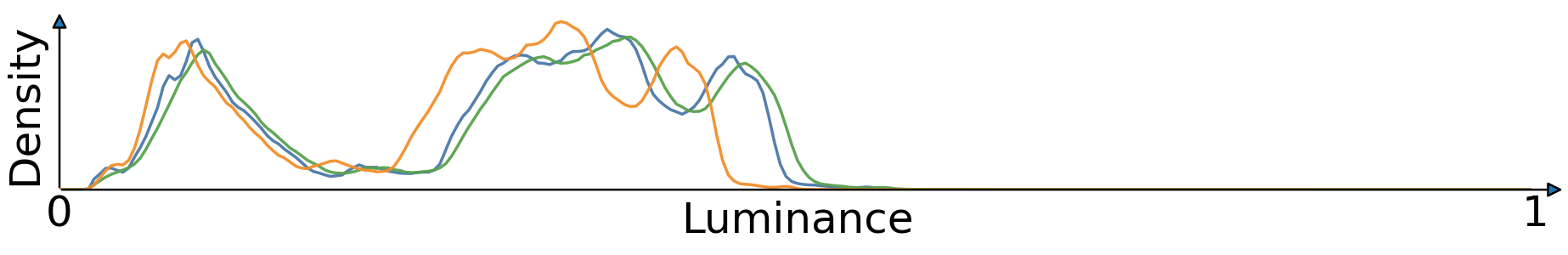}
    \vspace{-2em}
    \caption{}
    \label{fig:qualitative_color}
  \end{subfigure}
  \begin{subfigure}[h]{0.9\linewidth}
    \centering
    \includegraphics[width=0.92\linewidth]{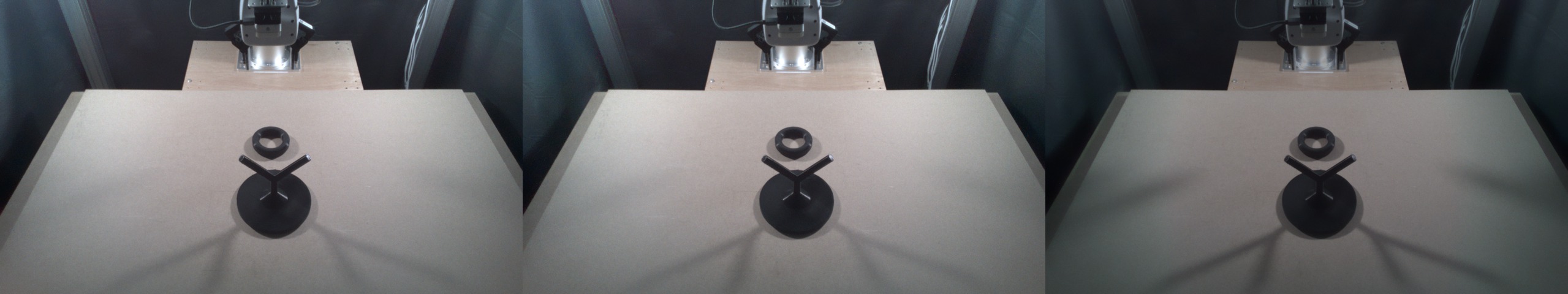}
    \includegraphics[width=\linewidth]{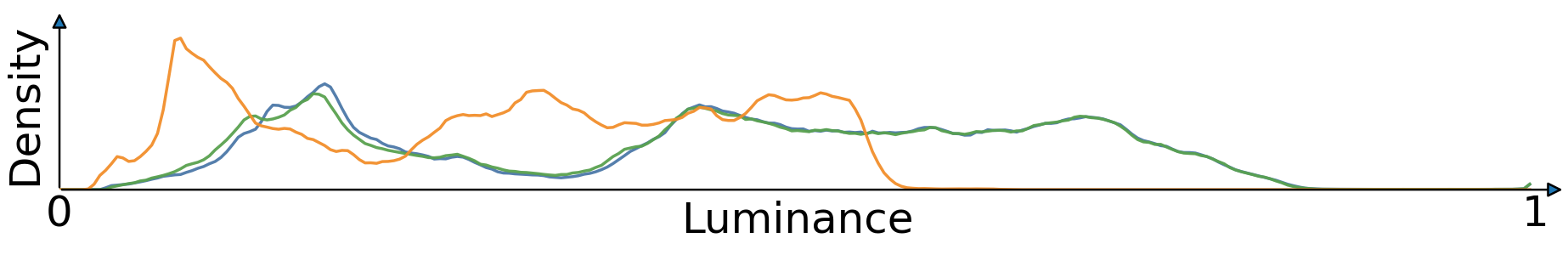}
    \vspace{-2em}
    \caption{}
    \label{fig:qualitative_direction}
  \end{subfigure}
  \begin{subfigure}[h]{0.9\linewidth}
    \centering
    \includegraphics[width=0.92\linewidth]{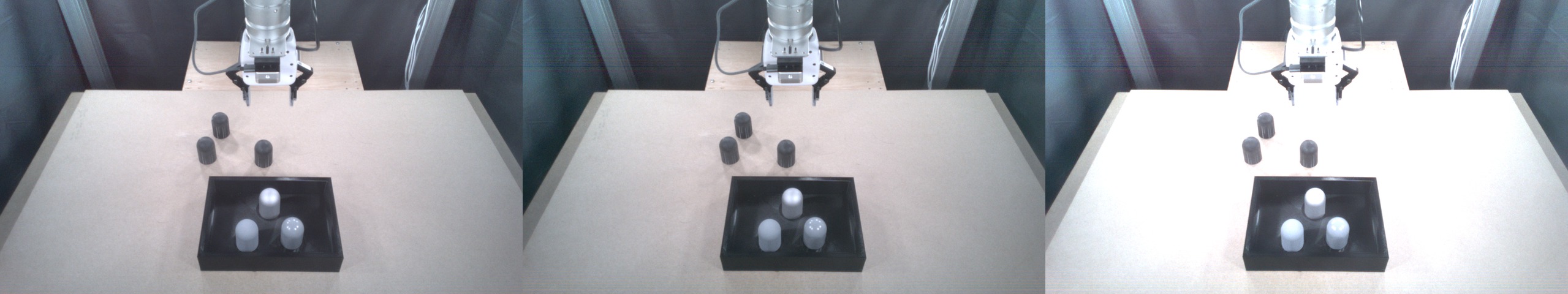}
    \includegraphics[width=\linewidth]{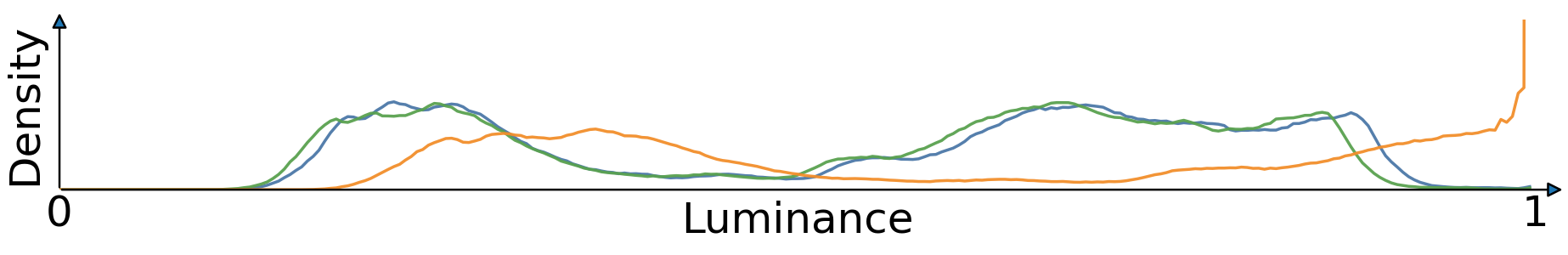}
    \vspace{-2em}
    \caption{}
    \label{fig:qualitative_intensity}
  \end{subfigure}
  
{\footnotesize
\textcolor[rgb]{0.30,0.47,0.66}{\rule{1em}{0.8pt}} Ground Truth \quad
\textcolor[rgb]{0.35,0.63,0.31}{\rule{1em}{0.8pt}}\textit{ RoboLight-Synthetic} \quad
\textcolor[rgb]{0.95,0.56,0.17}{\rule{1em}{0.8pt}} PNG-Average
}
    \caption{Visual fidelity verification. From left to right, we compare the ground-truth frame, the synthesized frame from \textit{RoboLight-Synthetic}, and frames generated by averaging PNG images across (a) color, (b) direction, and (c) intensity. For each comparison, the luminance histogram is included to provide a quantitative view of the luminance distribution.}
    \label{fig:qualtative}
\vspace{-1em}
\end{figure}

We evaluate the quality of \textit{RoboLight-Synthetic} from two perspectives: visual fidelity and effectiveness in policy training. \textit{Fig.}~\ref{fig:qualtative} compares the ground-truth frame, the synthesized frame from \textit{RoboLight-Synthetic}, and frames generated by averaging PNG images within the color, direction, and intensity subsets. 
The corresponding luminance histograms are included to provide a quantitative characterization of the illumination distribution.
The synthesized \textit{RoboLight-Synthetic} frames closely match the ground-truth frame, reflecting the radiometric accuracy preserved in HDR space.
Averaging LDR PNG images introduces deviations from the ground-truth frame, due to the limited illumination precision and non-linearity of LDR representations.

We further evaluate the effectiveness of \textit{RoboLight-Synthetic} for policy training by training Diffusion Policy from scratch using synthesized data and comparing it against a policy trained with data collected under the corresponding ground-truth lighting. 
Evaluation is conducted using 20 roll-outs per task. 
As reported in \textit{Tab.}~\ref{tab:quantative}, the results are comparable, with minor differences that may stem from small discrepancies between the ground-truth and synthesized lighting, as indicated by the luminance histogram in \textit{Fig.}~\ref{fig:qualtative}.
\renewcommand{\arraystretch}{1.2} 
\setlength{\tabcolsep}{1pt} 
\begin{table}[h]
\vspace{-1em}
\centering
\caption{Policy training effectiveness evaluation.}
\label{tab:quantative}
\begin{tabular}{lccc} 
\toprule
&\textbf{RGB Stacking} & \textbf{Donut Hanging} & \textbf{Sparkling Sorting} \\ 
\midrule
Ground Truth & 0.70 & 0.95& 0.35\\
\textit{RoboLight-Synthetic} & 0.65 & 0.80& 0.25\\
\bottomrule
\end{tabular}
\begin{tablenotes}
    \item Experimental settings follow those described in \textit{Tab.}~\ref{tab:task_difficulty}.
\end{tablenotes}
\vspace{-1em}
\end{table}

\section{Dataset Usage}
\subsection{A Lighting Robustness Benchmark}
\begin{figure}[h]
\vspace{-1em}
\centering
  \begin{subfigure}[t]{0.32\linewidth}
    \centering
    \includegraphics[width=\linewidth]{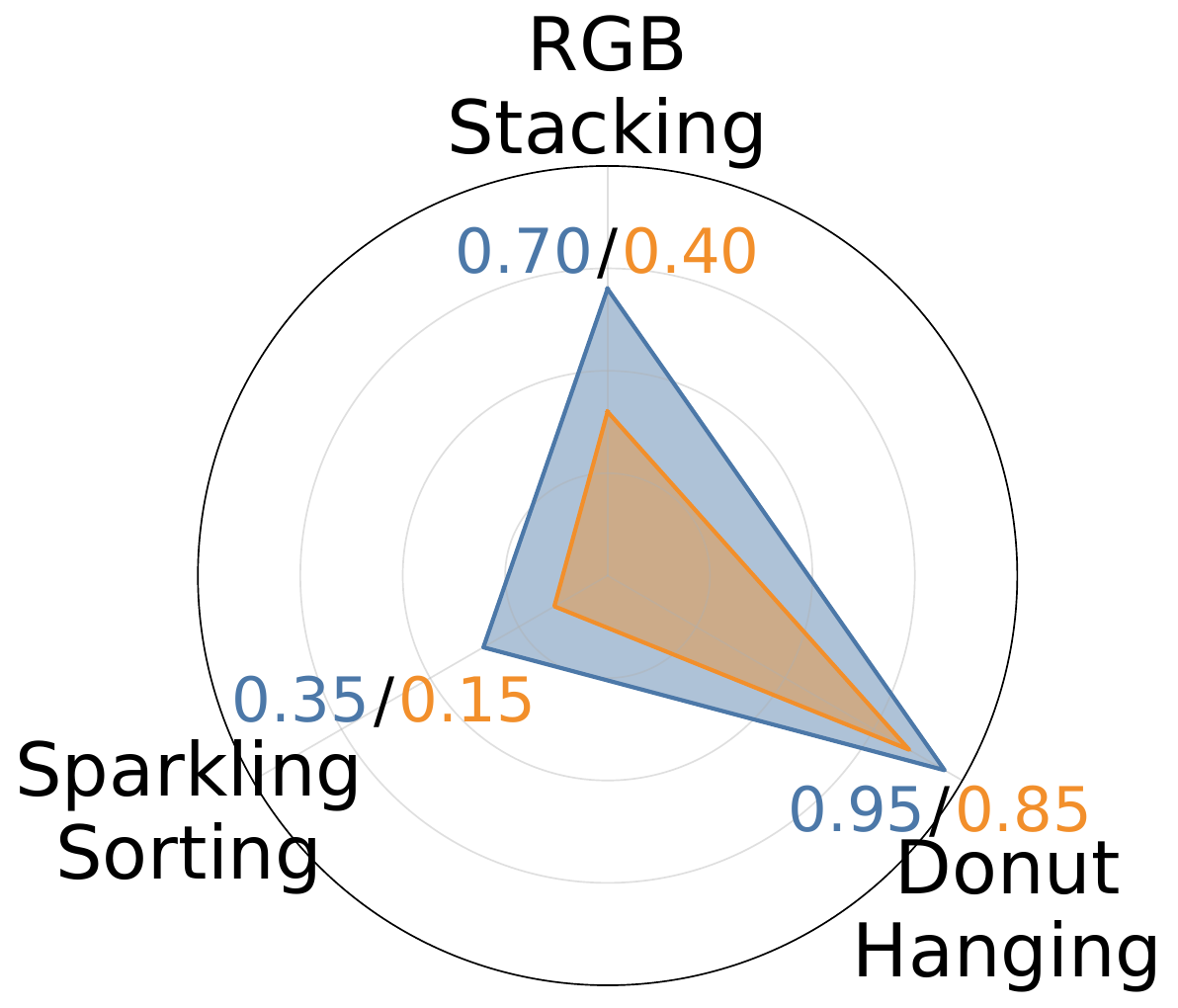}
    \vspace{-1em}
    \caption{Color}
    \label{fig:qualitative_color}
  \end{subfigure}
  \begin{subfigure}[t]{0.32\linewidth}
    \centering
    \includegraphics[width=\linewidth]{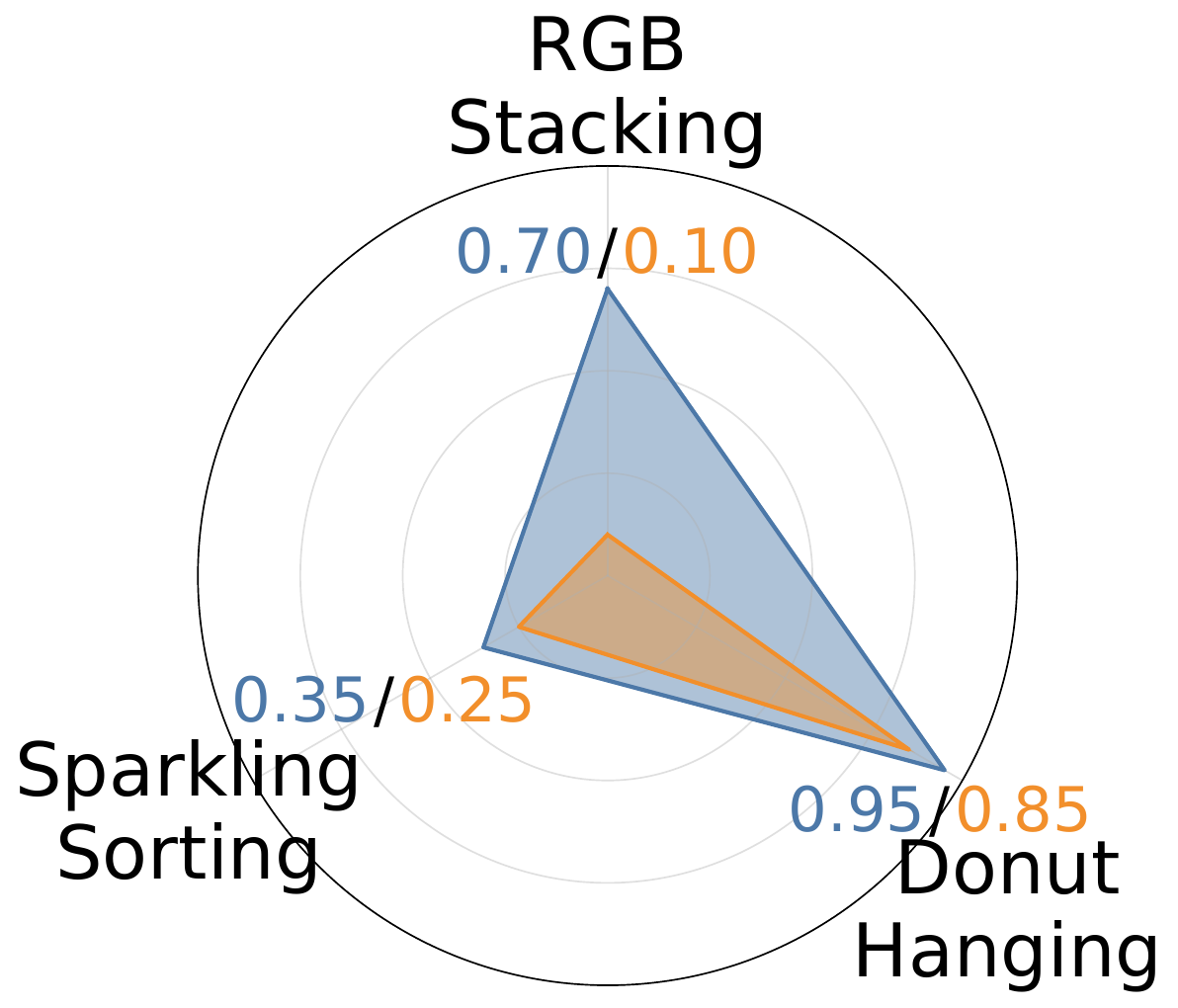}
    \vspace{-1em}
    \caption{Direction}
    \label{fig:qualitative_direction}
  \end{subfigure}
  \begin{subfigure}[t]{0.32\linewidth}
    \centering
    \includegraphics[width=\linewidth]{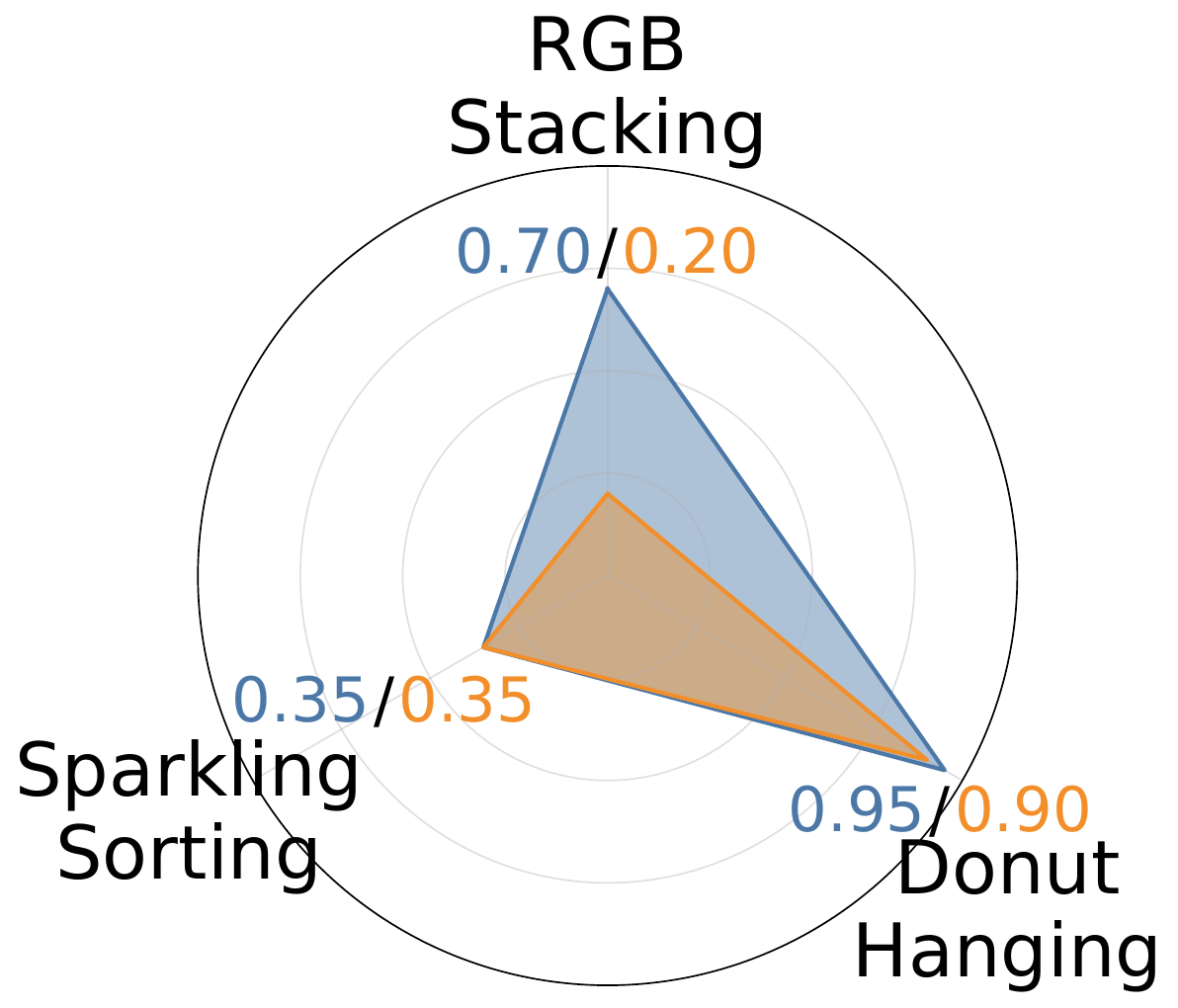}
    \vspace{-1em}
    \caption{Intensity}
    \label{fig:qualitative_intensity}
    \vspace{0.3em}
  \end{subfigure}
{\footnotesize
\textcolor[rgb]{0.30,0.47,0.66}{\rule{1em}{0.8pt}} Reference Lighting \quad
\textcolor[rgb]{0.95,0.56,0.17}{\rule{1em}{0.8pt}} Shifted Lighting
}
    \caption{Comparison of Diffusion Policy success rates under the reference lighting and under lighting shifted in (a) color, (b) direction, and (c) intensity across three tasks. The reference lighting setup is defined in \textit{Tab.}~\ref{tab:task_difficulty}.}
\label{fig:radar_light_shift}
\end{figure}

Policy robustness to viewpoint and texture shifts has been extensively studied and guides future dataset design in robotics~\cite{xie2024decomposing}.
With the precise lighting control of the \textit{Light Cube} and the \textit{RoboLight} dataset, we evaluate the robustness of Diffusion Policy to lighting shifts.
Three lighting shift conditions are defined:
(a) color variation, where the blue channel value in the lighting RGB configuration is reduced by 50\% relative to the training setup;
(b) direction variation, where illumination from the right-side lights is reduced by 50\%; 
and
(c) intensity variation, where overall light intensity is reduced by 50\%,  all relative to the lighting configuration in the training dataset.
As shown in \textit{Fig.}~\ref{fig:radar_light_shift}, lighting color shifts lead to the most consistent performance drop across the three tasks, compared to direction and intensity shifts. 
Among the tasks, \textit{RGB Stacking} exhibits the highest sensitivity to lighting variation. 
During evaluation, we find most failures under lighting shifts stem from object mislocalization during grasping, where the gripper closes without securing the object and proceeds to stack with an empty grasp.
To further analyze the effect of lighting variation, we introduce additional evaluation metrics. We report stage-wise success rates (\textit{S. R. }) for the sequential task of stacking the red, green, and blue cubes, as well as prediction error (\textit{P. E. }), defined as the distance (in millimeters) between the predicted object position at gripper closure and the ground-truth position. 
As shown in \textit{Tab.}~\ref{tab:RGB_Stacking_performance}, prediction error correlates with stage success rates, suggesting that lighting primarily affects policy performance through object localization. 
We also observe lighting-dependent biases: color shifts disproportionately affect localization of certain colored objects (e.g., the green cube), while direction shifts produce more uniform degradation across objects. 
However, with only 20 roll-outs per task under each lighting shift (180 in total), these findings may be influenced by evaluation variability, such as differences in initial object configurations. 
We will extend this protocol to the full dataset and additional policy architectures to establish a broader lighting robustness benchmark and conduct a more rigorous analysis of lighting effects on robotic policies.

\renewcommand{\arraystretch}{1.2}
\setlength{\tabcolsep}{6pt}

\begin{table}[h]
\centering
\caption{Policy performance on the RGB Stacking task under lighting shifts, reporting stage-wise success rate (\textit{S. R. }, $\uparrow$) and prediction error (\textit{P. E. }, $\downarrow$).}
\begin{tabular}{lcc|cc|cc}
\toprule
& \multicolumn{2}{c}{\textbf{Red Cube}} 
& \multicolumn{2}{c}{\textbf{Green Cube}} 
& \multicolumn{2}{c}{\textbf{Blue Cube}} \\
\cmidrule(lr){2-3}
\cmidrule(lr){4-5}
\cmidrule(lr){6-7}
& \textit{S. R.} & \textit{P. E.}
& \textit{S. R.} & \textit{P. E.}
& \textit{S. R.} & \textit{P. E.} \\
\midrule
Reference &0.80  & 11.83 & 0.70 & 13.15 &0.70  & 5.96 \\
Color-Shifted & 0.80 & 15.62 &0.40 & 21.81 & 0.40 & 12.89 \\
Direction-Shifted & 0.55 & 26.89 &0.30  & 26.60 & 0.10 & 15.19 \\
Intensity-Shifted & 0.50 & 24.55 & 0.25 & 18.98 & 0.20 & 15.82 \\
\bottomrule
\end{tabular}
\label{tab:RGB_Stacking_performance}
\vspace{-1em}
\end{table}

\subsection{Lighting Estimation for Robotic Manipulation}
\begin{figure}[h]
    \centering
    \begin{subfigure}{0.355\linewidth}
        \centering
        \includegraphics[width=\linewidth]{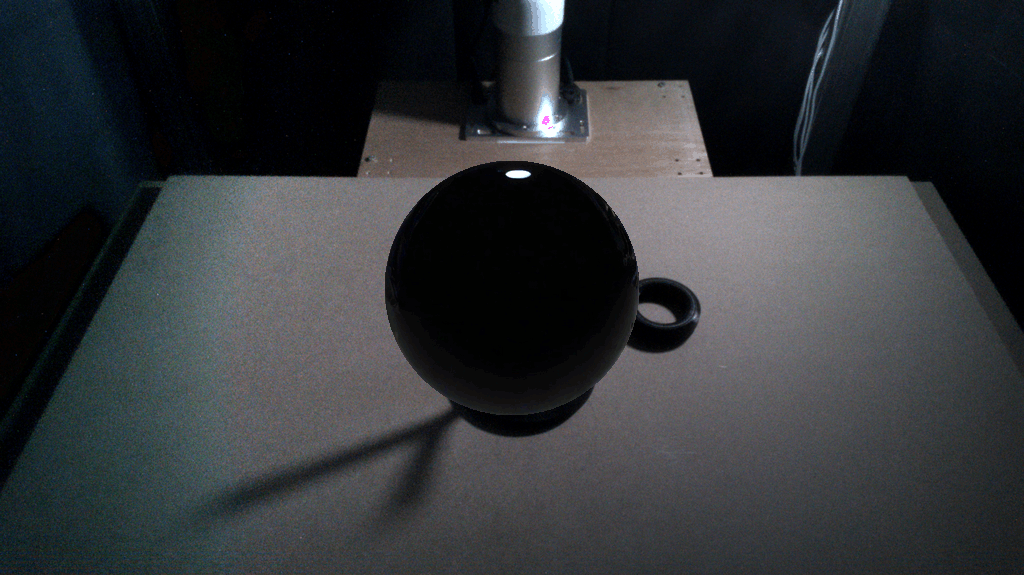}
        \vspace{-1.5em}
        \caption{}
    \end{subfigure}
    \begin{subfigure}{0.40\linewidth}
        \centering
        \includegraphics[width=\linewidth]{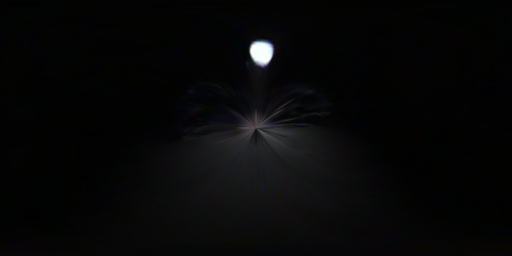}
        \vspace{-1.5em}
        \caption{}
    \end{subfigure}
    \caption{Lighting estimation results produced by DiffusionLight~\cite{phongthawee2024diffusionlight} on \textit{Donut Hanging} (Right lighting). (a) DiffusionLight estimates scene illumination by inpainting a virtual mirror ball at the image center, which reflects the surrounding environment and lighting. (b) The unwrapped environment map derived from the mirror ball, visualizing the predicted lighting color, direction, and intensity; the bright region indicates the estimated light source.}
    \label{fig:lighting_estimation}
    \vspace{-1em}
\end{figure}

Lighting estimation aims to recover scene illumination parameters from visual observations~\cite{phongthawee2024diffusionlight}. 
In robotics, these estimates could serve as illumination priors to guide data synthesis under previously unseen lighting conditions.
As illustrated in \textit{Fig.}~\ref{fig:lighting_estimation}, we apply DiffusionLight~\cite{phongthawee2024diffusionlight}, a zero-shot lighting estimation method, to scenes in our dataset. The estimated lighting direction can then be used to select or synthesize training data with corresponding illumination, enabling policy training tailored to the current environment.
For more complex lighting scenarios, this approach may be combined with illumination composition theory to construct datasets with richer lighting configurations~\cite{belhumeur1998set}.
The explicit lighting parameterization in \textit{RoboLight} offers a structured basis for investigating lighting estimation and composition in robotic manipulation.

\subsection{HDR-enabled Visual Condition Scaling}
\begin{figure}[h]
    \centering
    \begin{subfigure}{0.32\linewidth}
        \centering
        \includegraphics[width=\linewidth]{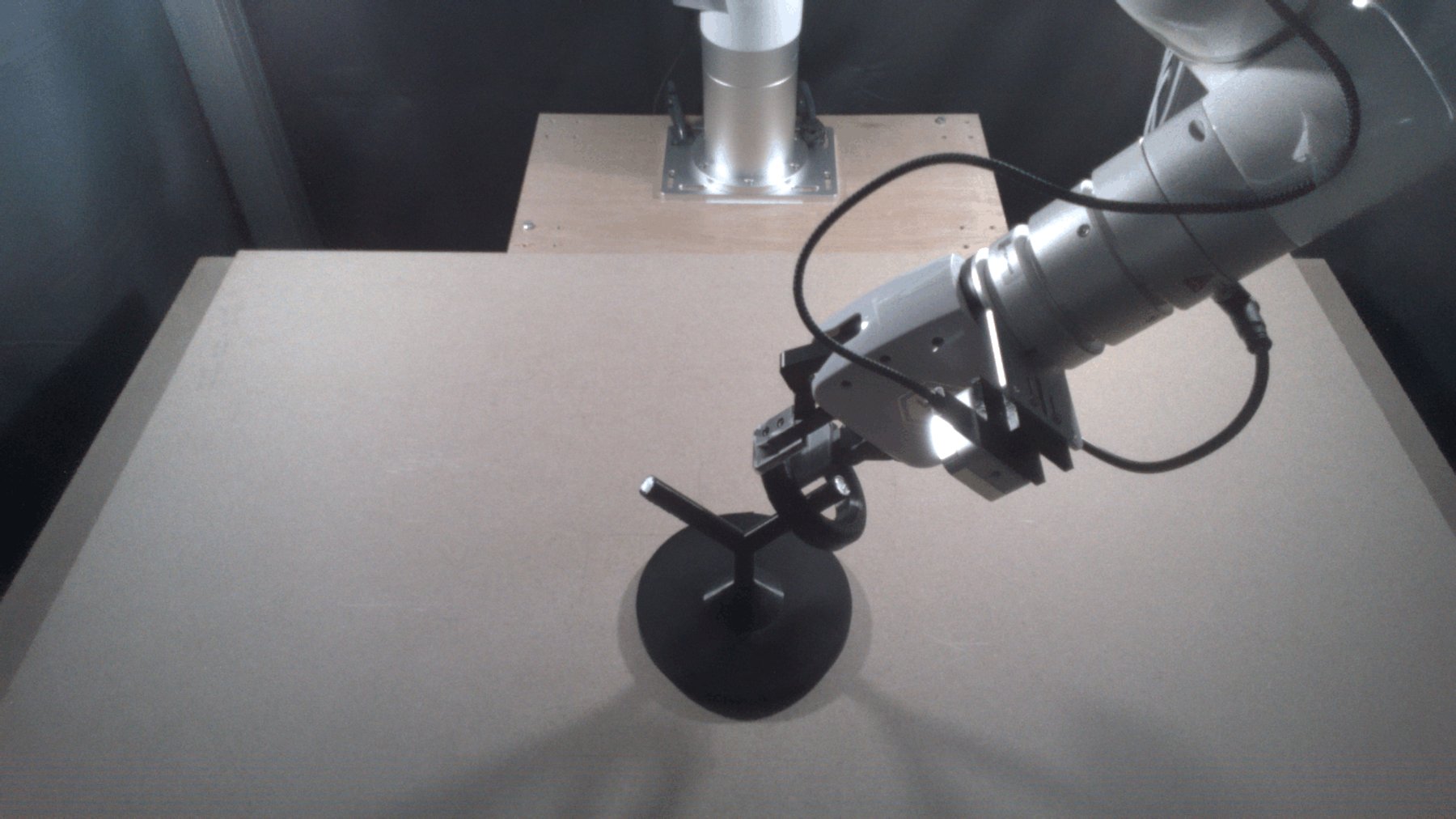}
        \vspace{-1.5em}
        \caption{}
    \end{subfigure}
    \begin{subfigure}{0.32\linewidth}
        \centering
        \includegraphics[width=\linewidth]{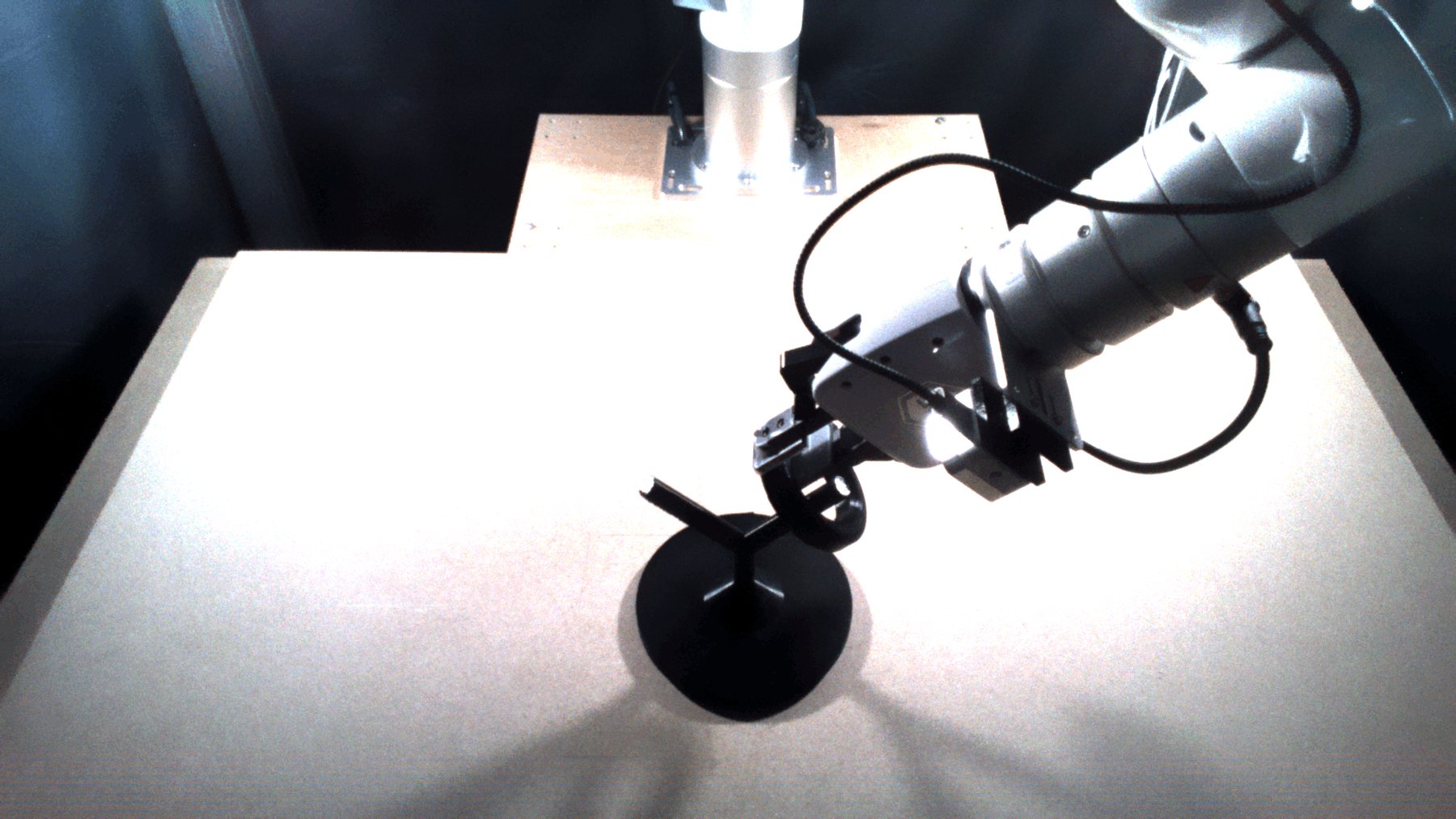}
        \vspace{-1.5em}
        \caption{}
    \end{subfigure}
    \begin{subfigure}{0.32\linewidth}
        \centering
        \includegraphics[width=\linewidth]{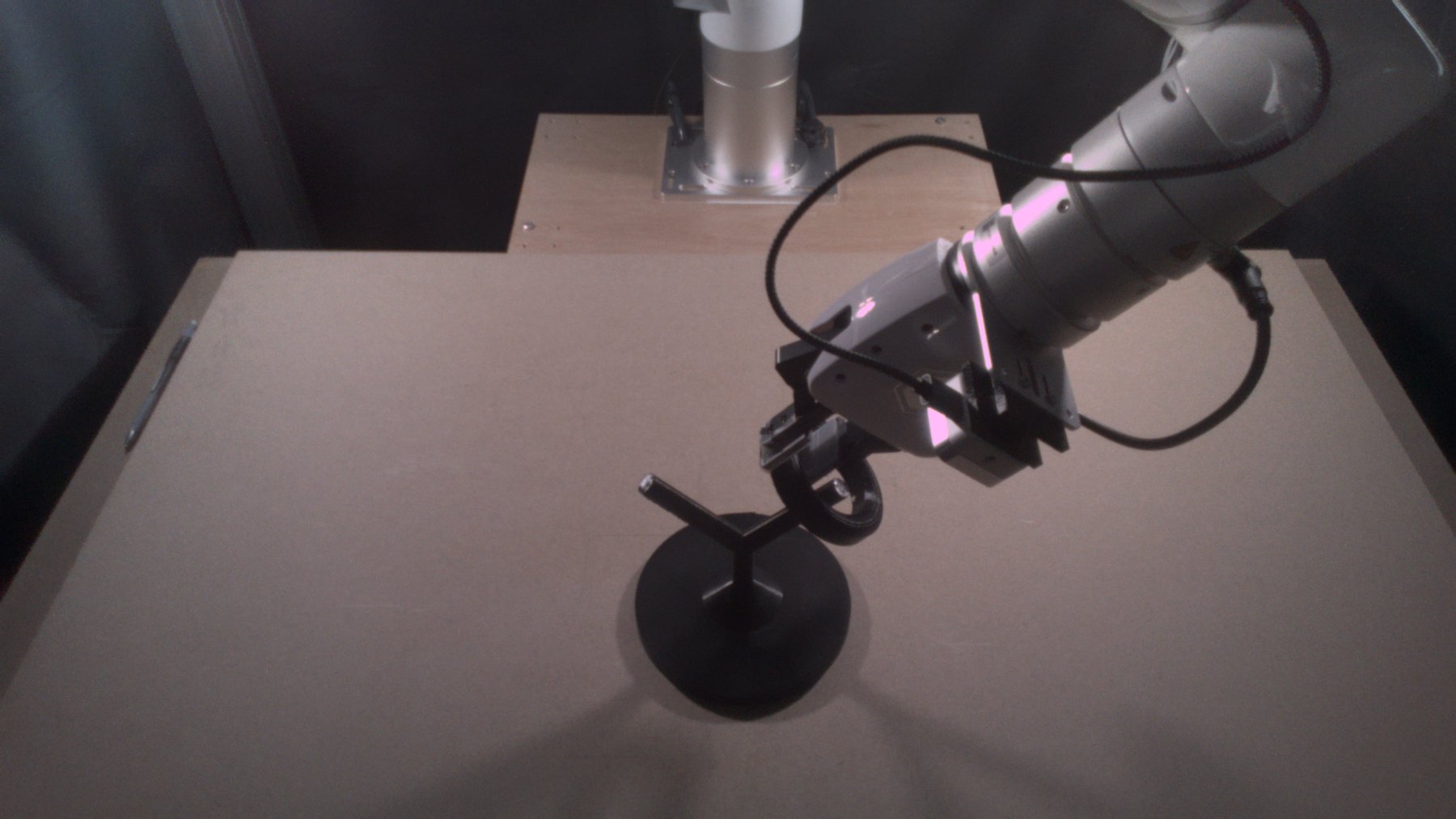}
        \vspace{-1.5em}
        \caption{}
    \end{subfigure}
    \caption{Examples of visual condition scaling via HDR post-processing: (a) ambient white lighting, (b) high-exposure conditions and (c) dim illumination.}
    \label{fig:hdr_enabled_visual_scaling}
\end{figure}

HDR image formats are widely adopted in photography due to their ability to preserve scene radiance over a high dynamic range, enabling flexible post-hoc exposure adjustment and tone mapping to produce varied global visual appearances.
Since our dataset is stored in HDR format, similar global radiometric transformations can be applied to generate diverse visual conditions.
Examples are shown in \textit{Fig.}~\ref{fig:hdr_enabled_visual_scaling}. By adjusting global exposure and tone mapping of the HDR images, we generate (a) ambient white lighting, (b) high-exposure conditions resembling industrial environments, and (c) dimmer, yellow-toned illumination. 
These global transformations can be consistently applied to all HDR episodes in \textit{RoboLight}, enabling scalable augmentation of visual conditions through post-hoc adjustments.

\section{Conclusion}
\paragraph{Limitations and Future Work}
A primary limitation lies in the procedure for collecting synchronized episodes, where we replay the same trajectory under varied lighting and rely on recorded sensor timestamps to preserve temporal alignment. 
However, this process remains sensitive to small deviations, such as slight object displacement during initialization or minor variations in grasp execution. 
These errors can accumulate along the trajectory, leading to desynchronization across episodes.
Manual inspection is therefore required to ensure consistency across lighting conditions, making the data collection process more labor-intensive than standard procedures.
Future work will explore replacing the current Bluetooth-controlled lighting system with a high-frequency illumination switching system driven by embedded controllers, enabling multiple lighting conditions to be captured in a single execution.

\paragraph{Conclusions}
We propose \textit{RoboLight}, the first dataset capturing synchronized episodes under systematically varied lighting conditions.
The dataset comprises two components: (a) \textit{RoboLight-Real}, a real-world dataset containing 2,800 episodes collected within the \textit{Light Cube}, which we designed for precise control of illumination; and (b) \textit{RoboLight-Synthetic}, a theoretically unbounded dataset synthesized based on the linearity of light transport.
All image data are recorded in HDR format for maximized radiometric accuracy.
We further verify dataset quality through qualitative analysis and real-world policy roll-outs, evaluating task difficulty, distributional diversity, and synthesized data effectiveness.
Additionally, we demonstrate three representative use cases of the proposed dataset, ranging from lighting robustness benchmarking to collection-free visual scaling enabled by the HDR format.
To support continued research, we open-source the full dataset, together with the system software for generating HDR frames using widely accessible Intel RealSense cameras. 
We also provide hardware design specifications for constructing the \textit{Light Cube} and object sets to support reproducibility and cross-group experimental use.

\bibliographystyle{IEEEtran}
\balance
\bibliography{citations}

\end{document}